\documentclass[11pt, a4paper, logo, copyright]{eai}
\usepackage{shlab}

\usepackage[numbers, sort&compress]{natbib}
\usepackage{amsmath,amsfonts,bm}
\usepackage{multirow}
\usepackage{subcaption}
\usepackage{wrapfig}

\def\eqref#1{equation~\ref{#1}}

\def\1{\bm{1}}

\DeclareMathAlphabet{\mathsfit}{\encodingdefault}{\sfdefault}{m}{sl}
\SetMathAlphabet{\mathsfit}{bold}{\encodingdefault}{\sfdefault}{bx}{n}

\usepackage{array}
\usepackage{booktabs}
\usepackage{tabularx}
\usepackage{multirow}
\usepackage{wrapfig}
\usepackage{ragged2e}
\usepackage{algorithm}
\usepackage{algpseudocode}
\usepackage{xspace}
\usepackage[most]{tcolorbox}
\usepackage{placeins}
\usepackage{cleveref}
\crefname{section}{Section}{Sections}
\Crefname{section}{Section}{Sections}
\crefname{subsection}{Section}{Sections}
\Crefname{subsection}{Section}{Sections}
\graphicspath{{}{appendix/}}

\definecolor{figblue}{RGB}{70,90,155}
\definecolor{figgreen}{RGB}{45,130,85}
\definecolor{figpink}{RGB}{195,65,72}
\definecolor{figpurple}{RGB}{130,70,148}
\definecolor{TeacherYellow}{RGB}{255,248,170}
\definecolor{promptheaderbg}{RGB}{50,50,55}
\definecolor{promptborder}{RGB}{180,180,185}
\definecolor{promptbodybg}{RGB}{252,252,252}

\newcommand{\eg}{e.g.,\xspace}

\DeclareRobustCommand{\realname}{\texorpdfstring{{\normalfont\sffamily\bfseries\textcolor{linkcolor}{REAL}}}{REAL}}
\DeclareRobustCommand{\real}{\realname\xspace}
\DeclareRobustCommand{\realbench}{\realname\mbox{-Bench}\xspace}
\DeclareRobustCommand{\realagent}{\realname\mbox{-Agent}\xspace}
\newcommand{\promptcardfile}[2][]{%
  \tcbinputlisting{%
    enhanced,
    breakable,
    colframe  = promptborder,
    colback   = promptbodybg,
    boxrule   = 0.5pt,
    arc       = 1.5pt,
    left      = 6pt, right = 6pt,
    top       = 24pt, bottom = 4pt,
    listing only,
    listing file={appendix/#2},
    listing options={
      basicstyle=\small\ttfamily,
      breaklines=true,
      breakautoindent=false,
      columns=fullflexible,
      aboveskip=0pt, belowskip=0pt,
    },
    attach boxed title to top left = {yshift=-\tcboxedtitleheight},
    boxed title style = {
      colback  = promptheaderbg,
      colframe = promptheaderbg,
      arc      = 1.5pt,
      boxrule  = 0pt,
      left     = 6pt, right = 6pt,
      top      = 4pt, bottom = 4pt,
    },
    coltitle = white,
    fonttitle = \small\sffamily\bfseries,
    #1%
  }%
}

\title{Exploratory, Communicative, and Deployable: Vision-Driven Embodied Agents for Open-World Mobile Manipulation}

\newcommand{\equalcontrib}{\textsuperscript{*}}
\newcommand{\corrauth}{\textsuperscript{\faEnvelope}}
\newcommand{\affmark}[1]{\textsuperscript{#1}}

\makeatletter
\renewcommand{\@author}{%
\begin{tabular}{c}
Boyu Mi\affmark{*,1,2}\quad Mengchen Ma\affmark{*,2,3}\quad Yifei Yao\affmark{*,1,2}\quad Xing Gao\affmark{2}\corrauth\\
Junting Chen\affmark{4}\quad Yangzi Li\affmark{5}\quad Zihou Zhu\affmark{2}\quad Guohao Li\affmark{6}\quad Zhenfei Yin\affmark{7}\\
Tai Wang\affmark{2}\quad Yao Mu\affmark{1,2}\quad Jiangmiao Pang\affmark{2}\quad Hanqing Wang\affmark{2}\corrauth
\end{tabular}\\[0.35em]
\begin{tabular}{c}
\affmark{1}Shanghai Jiao Tong University\quad
\affmark{2}Shanghai Artificial Intelligence Laboratory\\
\affmark{3}Southeast University\quad
\affmark{4}National University of Singapore\quad
\affmark{5}Zhejiang University\\
\affmark{6}CAMEL-AI.org\quad
\affmark{7}University of Oxford
\end{tabular}\\[0.25em]
{\fontsize{8}{10}\selectfont \equalcontrib\ Equal contribution.}\\[-0.05em]
{\fontsize{8}{10}\selectfont \corrauth\ Corresponding authors: Xing Gao (\texttt{gaoxing@pjlab.org.cn}); Hanqing Wang (\texttt{hanqingwang.c@gmail.com}).}
}
\makeatother


\begin{document}

\begin{abstract}
Real-world deployment of embodied agents requires active exploration, visual grounding, and interactive intent disambiguation. However, existing frameworks often rely on privileged simulator states or assume complete instructions, bypassing realistic deployment challenges. To bridge this gap, we present \real, an agentic framework for open-world mobile manipulation. \real establishes sim-to-real-consistent environment APIs without oracle perception and integrates a simulated user to enable human-in-the-loop interaction. Within this environment, we design diverse task compositions to drive data collection, supervised fine-tuning, and online reinforcement learning, systematically optimizing agent performance. To comprehensively evaluate this approach, we introduce \realbench, a benchmark spanning 241 tasks across active exploration, visual distraction, articulated manipulation, and interactive disambiguation. 
Experimental results demonstrate that our trained agent outperforms leading commercial closed-source VLMs on interactive tasks with a 56.9\% success rate. Further empirical analysis reveals that our hierarchical training pipeline successfully aligns the model's tool-use capabilities while maintaining robust open-vocabulary reasoning under extended exploration horizons. Finally, we deploy and evaluate our framework on a physical dual-arm mobile robot, where it achieves a 78.3\% end-to-end success rate over 60 real-world episodes. These physical trials demonstrate robust zero-shot transferability to unseen household scenarios, validating that our sim-to-real-consistent design successfully bridges the reality gap for long-horizon mobile manipulation. Code is available at \href{https://github.com/InternRobotics/REAL}{\textnormal{github.com/InternRobotics/REAL}}.

\end{abstract}

\maketitle

\section{Introduction}

\begin{figure*}[t]
  \centering
  \includegraphics[width=\linewidth]{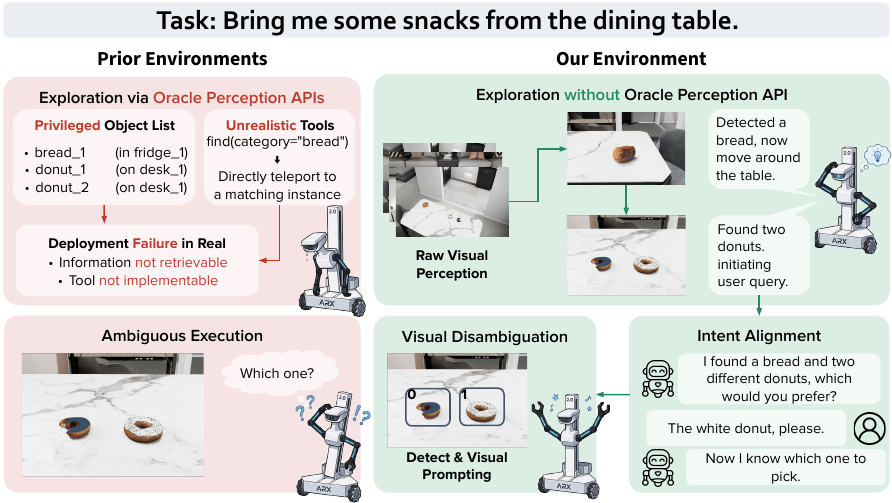}
  \caption{\textbf{Comparison of embodied agent frameworks.} \textbf{Left:} Reliance on oracle perceptual APIs and open-loop execution in prior environments hinders ambiguity resolution and real-world deployment. \textbf{Right:} \real removes oracle perceptual APIs while retaining physically obtainable scene priors, enabling RGB-based exploration and natural language interaction for ambiguous instructions.}
  \label{fig:head}
\end{figure*}

With their advanced reasoning and planning capabilities, Large Language Models (LLMs) have become a primary driver for embodied agents~\cite{ichter2022saycan,huang2022language,huang2022inner,liang2023code,szot2023llmpolicies,li2024embodiedinterface}. Early frameworks primarily utilized these models to execute diverse household tasks within text-based environments built upon underlying physical simulators~\cite{kolve2017ai2thor,szot2021habitat,shridhar2021alfworld,puig2018virtualhome,li2023behavior,shridhar2020alfred} via textual perception and tool calling. However, pure-text interactions inherently lack the spatial and physical grounding critical for complex execution. This limitation has driven the widespread adoption of Vision-Language Models (VLMs) to endow agents with rich visual perception~\cite{brohan2022rt1,szot2025multimodal,zitkovich2023rt2,driess2023palme}. By leveraging high-fidelity rendered imagery, modern vision-driven environments enable precise spatial reasoning and significantly expand the range of executable tasks~\cite{wang2025votabench,yang2025embodiedbench,dai2025manitaskgen}. Yet, a critical bottleneck persists: existing benchmarks often decouple simulation from real-world complexities across two fundamental dimensions—active exploration and communicative interaction (see \cref{fig:head}, Left). Bypassing these realistic constraints inherently leads to deployment failures during sim-to-real transfer.

Specifically, the first limitation lies in the reliance on simulator-privileged perceptual information during physical exploration. Many existing environments equip agents with idealized tools that access global object lists~\cite{li2024embodiedinterface,dai2025manitaskgen,yang2025embodiedbench} or locate targets via oracle category-finding APIs~\cite{yang2025embodiedbench, wang2025votabench}. Since these privileged interfaces are difficult to retrieve in physical environments, policies trained within such setups can face performance discrepancies during real-world deployment. To address this, several recent works~\cite{chang2024partnr, li2025emmoe, ren2024explore} adopt a non-privileged paradigm that restricts agents to standard sensor inputs like RGB images. However, their exploration and planning trajectories often operate primarily at the textual level rather than being tightly grounded in visual observations. Consequently, these methods can still struggle with precise visual disambiguation when encountering visually similar objects or distractors.

The second limitation stems from the assumption of explicit user instructions, leaving the need for dynamic intent alignment less addressed. Current benchmarks~\cite{chen2025eratransformingvlmsembodied,wang2025votabench,li2025emmoe} typically treat execution as an open-loop process, assuming that the user possesses sufficient knowledge of the environment and issues unambiguous commands. In practical deployment, however, human users often operate under incomplete information, leading to initially ambiguous goals that require active, context-aware interaction to clarify intent. While recent studies~\cite{patel2025adapt,wang2025famer,philipov2024simulating} have introduced simulated users to facilitate interaction, their scope is primarily focused on text-based preference elicitation or memorization, leaving the co-optimization of communication and physical mobile manipulation tasks less explored.

To resolve these issues, we present \real (explo\textbf{R}atory, communicativ\textbf{E}, \textbf{A}nd, deploy\textbf{L}able), an agentic framework designed for interactive and deployable open-world mobile manipulation. As illustrated in \cref{fig:head}~(Right), \real establishes a closed-loop visual environment entirely free of oracle perceptual APIs. The agent is restricted from accessing global object lists, target 3D poses, privileged simulator states, or goal locations. Instead, to ground objects, the agent must actively explore the environment and discover objects from raw RGB observations through a deployable toolchain spanning navigation, object detection, and visual prompting for manipulation~\cite{nasiriany2024pivot}. In addition, we integrate a simulated user that issues initially ambiguous instructions and provides dynamic, context-aware natural language feedback. This feedback is tightly grounded in the agent's ongoing exploration results, effectively capturing the information blind spots inherent in real-world human interactions.

Within this environment, we develop an automated, scalable data generation and filtering pipeline to construct diverse exploration and interaction tasks. 
Utilizing the resulting trajectory data, we align and optimize the \real agent based on Qwen3-VL-8B-Instruct~\cite{qwen2025qwen25} via a two-stage paradigm: supervised fine-tuning (SFT) for tool-use alignment, followed by online reinforcement learning (RL) to enhance adaptive reasoning and error recovery.
To systematically evaluate our agent, we introduce \realbench, a comprehensive benchmark comprising 241 task instances across four distinct families: active exploration, visual distraction, articulated manipulation, and interactive disambiguation. 
Experimental results show that our trained agent successfully outperforms leading commercial closed-source VLMs, achieving a 56.9\% success rate on interactive tasks. Finally, we deploy and evaluate our agent on a physical ARX LIFT2 dual-arm mobile robot. The high-level policy achieves a 78.3\% end-to-end success rate over 60 real-world episodes, demonstrating robust zero-shot transferability and validating that our sim-to-real-consistent framework design effectively bridges the reality gap.

In summary, our main contributions are threefold:
\begin{enumerate}
    \item \textbf{\realname{} framework.} We introduce \real, a sim-to-real-consistent agentic framework without oracle perception, coupled with a simulated user for dynamic human-in-the-loop intent alignment.
    \item \textbf{Training pipeline.} We develop an automated task-generation pipeline and train a Qwen3-VL-8B-based agent with dual-stage SFT and online RL, enabling robust tool-use alignment and adaptive replanning.
    \item \textbf{Benchmark and deployment.} We construct \realbench with 241 tasks and show that the trained agent outperforms strong commercial baselines in simulation while transferring zero-shot to a physical dual-arm mobile robot with a 78.3\% success rate in real-world tasks.
\end{enumerate}

\section{Related Work}

\subsection{Simulation Environments and Benchmarks for Embodied Agents}
Many embodied-agent benchmarks are instantiated in foundational simulation platforms and task environments~\cite{kolve2017ai2thor, szot2021habitat, shridhar2021alfworld, puig2018virtualhome, li2023behavior, lei2025embomatrix, deitke2022procthor, yang2023holodeck}. Beyond task coverage, these benchmarks differ in the policy-facing observations and action interfaces they expose, which directly affects sim-to-real transfer. The Embodied Agent Interface~\cite{li2024embodiedinterface} and ManiTaskGen~\cite{dai2025manitaskgen} evaluate LLMs or generate fine-grained tasks, but simplify exploration by providing ground-truth object metadata. VisualAgentBench~\cite{liu2025visualagentbench}, EmbodiedBench~\cite{yang2025embodiedbench}, and VoTa-Bench~\cite{wang2025votabench} incorporate visual observations, but some evaluation interfaces expose simulator-specific or oracle-assisted primitives, such as category-level ``find,'' that are difficult to realize reliably on physical robots. PARTNR~\cite{chang2024partnr} supports room exploration through Concept Graphs~\cite{gu2024conceptgraphs}, while EMMOE~\cite{li2025emmoe} combines a high-level planner with non-privileged low-level policies; nevertheless, their high-level interfaces refer to discovered objects primarily through text, limiting instance-level disambiguation among visually similar distractors. Some works contribute both a benchmark and an accompanying agent; here we discuss their evaluation interfaces, while the next subsection focuses on their policy designs. In contrast, our environment and \realbench ground discovered objects in visual space and evaluate active exploration without exposing global object lists, target poses, or oracle category-search APIs.

\subsection{High-Level Embodied Agent Frameworks}
Vision-language embodied systems operate at different levels of abstraction. Low-level VLA policies, such as OpenVLA~\cite{kim2024openvla} and $\pi_{0.5}$~\cite{intelligence2025pi05}, map visual observations and language instructions to robot actions. They are execution policies rather than direct counterparts to the high-level agent studied here; in our physical system, a VLA model implements manipulation primitives beneath \real. At a higher level, LLM- and VLM-based agents reason over observations and orchestrate tools or reusable skills. Voyager~\cite{wang2023voyager} demonstrates open-ended exploration with an LLM agent. In simulation-oriented mobile-manipulation settings, VoTa-Bench~\cite{wang2025votabench} applies dual preference optimization to state prediction and action selection; EMMOE~\cite{li2025emmoe} trains a DPO-based planner above lightweight policies~\cite{chi2023diffusionpolicy}; ERA~\cite{chen2025eratransformingvlmsembodied} distills embodied priors and refines its agent through online RL; and OWMM-Agent~\cite{chen2025owmmagent} adapts VLMs through simulation-based multimodal agentic data synthesis. RoboOS~\cite{tan2025roboos} and Being-0~\cite{yuan2025being} further demonstrate modular embodied systems on physical robots. The distinguishing focus of \real is to train a high-level visual-interactive policy that unifies oracle-free active exploration, visually grounded object selection, and user interaction, while delegating physical execution to deployable low-level skills.

Classical task and motion planning (TAMP) integrates symbolic task planning with continuous motion feasibility and state tracking~\cite{kaelbling2011tamp,garrett2021tamp}. 
LLM planners such as SayCan~\cite{ichter2022saycan} ground language in predefined skills and affordances, while uncertainty-aware or replanning frameworks such as KnowNo~\cite{ren2023knowno} and RePLan~\cite{skreta2024replan} decide when to ask for help or recover from execution failures. \real is complementary to this line of work: it does not solve the TAMP problem or model all uncertainty in low-level motion execution. Instead, \real studies how to train a deployable visual-interactive policy that acquires missing task state through RGB exploration and user dialogue before dispatching physical skills.

\subsection{Interactive Embodied Agents and User Modeling}
TEACh~\cite{padmakumar2022teach} establishes multi-turn human-agent dialogue for household tasks, while VL-LN~\cite{huang2025vlln} uses simulated users to resolve ambiguities through active communication. For user intent understanding, ADAPT~\cite{patel2025adapt} actively elicits preferences through questions, and FARMER~\cite{wang2025famer} models unstated goals through theory of mind. Our focus is complementary: \real grounds simulated-user feedback in observations acquired during exploration, requiring the agent to alternate among visual perception, multi-turn dialogue, and manipulation to acquire missing task state and complete long-horizon tasks.

\section{Environment Design}
\label{sec:environment}

\begin{figure*}[t]
  \centering
  \includegraphics[width=\linewidth]{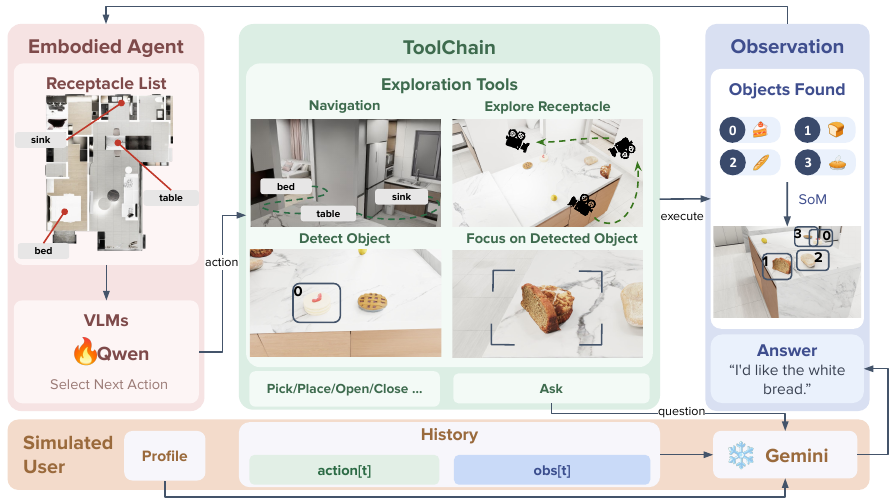}
  \caption{
  Overview of the environment design. The Embodied Agent receives physically obtainable receptacle priors and RGB/SoM observations with visual IDs, but no oracle object lists, poses, simulator states, or goal locations. The ToolChain exposes deployable exploration tools (\texttt{nav\_to}, \texttt{walk\_around}, \texttt{show\_object\_by\_category}, \texttt{gaze\_at}), \texttt{Pick/Place/Open/Close} controls, and \texttt{Ask}. The simulated user conditions on the agent's action/observation history and a task profile to answer queries; backend state may be used by handlers but is never exposed to the policy.}
  \label{fig:env}
\end{figure*}

To facilitate the deployment of vision-driven embodied agents from simulation to the real world, we construct a closed-loop embodied environment (\cref{fig:env}). Built upon the GRUtopia~\cite{wang2024grutopia} platform, the proposed environment features high-fidelity rendering and rigid-body physics. At the macro level, the design adheres to two core principles that distinguish it from previous environments that simplify exploration and communication: providing a toolchain free of oracle perceptual APIs and capable of real-world implementation, and integrating a simulated user for natural language interaction. Within this architecture, scene entities are categorized into fixed macroscopic \texttt{Receptacles} (\eg tables, cabinets) and manipulable \texttt{Objects} (\eg cups, books). For a specific scene, the \texttt{Receptacles} remain static across different task instances, whereas the quantities, categories, and spatial placements of manipulable \texttt{Objects} vary.

\subsection{Environment Modeling without Oracle Perception}
\label{subsec:privilege_free}

\paragraph{Initial Observation.}
We assume that the agent is equipped with a long-term 3D occupancy map and semantic segmentation of the receptacles, enabling navigation to a receptacle through path planning. This prior is physically obtainable from a robot's previous scans or mapping stack. Consequently, at task initialization, the agent receives only a prior list of receptacle IDs and their semantic categories (\eg \texttt{table\_2}; see the Embodied Agent block in \cref{fig:env}). Despite this structural prior, the agent has no access to global object lists, target 3D poses, simulator object states, or goal locations. It also lacks the fine-grained appearance of receptacles and the distribution of objects upon or within them. To ground target objects, the agent must employ the active exploration toolchain described below to acquire observations.

\paragraph{Multi-level Active Exploration.} 
To enable exploration, we introduce a multi-level active exploration toolchain (see the Exploration Tools block in \cref{fig:env}). These tools avoid oracle perceptual APIs at the policy interface and can be transferred to the real world (implementation details are provided in Appendix~\ref{sec:supp_tool_impl_per}). The overall process is structured into four sequential actions. The inter-receptacle navigation tool (\texttt{nav\_to}) enables movement between different receptacle nodes based on structural priors. Subsequently, the intra-receptacle scanning tool (\texttt{walk\_around}) functions as a broad-spectrum scanner driven by open-vocabulary 2D perception models. This tool navigates the agent around a specific receptacle to bypass physical occlusions and detect common object categories from multiple viewpoints. Following this, the object detection tool (\texttt{show\_object\_by\_category}) actively isolates objects of a user-defined category within the current ego-centric view. Ultimately, the target approach and alignment tool (\texttt{gaze\_at}) advances the agent toward a selected target to center it within the view of the camera, while accounting for physical distance constraints.

\paragraph{Vision-Grounded Interaction.}
To maintain spatial memory without relying on privileged 3D coordinates, the detected objects are dynamically back-projected from the 2D segmentation masks into the 3D space. This geometric projection constructs a temporary exploration map that records all discovered object categories and their spatial anchors during the most recent invocation of the exploration tool. Objects recorded in this map are highlighted within the RGB/SoM observation of the agent~\cite{yang2023setofmark} (see the Observation block in \cref{fig:env}). This mechanism bridges perception with physical interaction, enabling the agent to distinguish objects of the same category and parameterize downstream manipulation skills (\eg \texttt{pick}, \texttt{open}; see the \texttt{Pick/Place/Open/Close} of the ToolChain block in \cref{fig:env}) by referencing the assigned visual IDs.

\paragraph{Simulated User for Collaborative Tasks.}

To simulate collaborative tasks, we integrate a simulated user server driven by \texttt{gemini-3-flash}~\cite{gemini2025} directly into the dynamics of the environment (see the Simulated User block in \cref{fig:env}). At the top level, this module functions as an interactive supervisor that directs the overall task flow. It provides ambiguous initial instructions, monitors the task progress of the agent remotely, and progressively clarifies objectives by answering queries based on real-time exploration results.

To operationalize this top-level design, the module relies on specific mechanisms for configuration, synchronization, and communication. A task-specific profile is initialized for the simulated user to establish the boundaries of preferences and choices, such as restricting targets to objects currently present in the scene. This configuration ensures evaluation stability and prevents the generation of instructions involving non-existent or unexplored items. Furthermore, the server maintains continuous synchronization with the execution trajectory of the embodied agent. By accessing the interaction history and physical observations of the agent, the system constructs a consistent shared context for multi-turn alignment, accurately replicating a human user with limited perception of the initial scene layout. Finally, a bilateral communication interface facilitates interaction. When physical exploration alone fails to resolve ambiguity, the agent can issue an \texttt{ask} action to query the user (see the \texttt{Ask} block and question flow in \cref{fig:env}). Leveraging the shared context and predefined profile, the simulated user subsequently generates natural language guidance to clarify intent and drive the completion of the task.

\section{Agent Training}
\label{sec:method}

To minimize online inference overhead and mitigate risks of context explosion or temporal hallucinations during long-horizon deployment, we formulate the decision-making process as a Partially Observable Markov Decision Process (POMDP). Instead of continuously concatenating raw historical observations, our policy relies on a compact memory paired with the current visual input to derive discrete tool actions, maintaining a computationally efficient, localized historical proxy.

Our training pipeline consists of three steps. First, we construct a causal, closed-loop policy context and transition pipeline to support efficient data generation (\cref{sec:data_pipeline}). Second, we use expert demonstrations to align the model with the tool interface and structured reasoning format (\cref{sec:sft}). Finally, we optimize the policy with Group Sequence Policy Optimization (GSPO)~\cite{zheng2025group}, using exploration and environmental feedback to improve task-solving performance (\cref{sec:rl}).

\subsection{Problem Formulation and Context Management}
\label{sec:data_pipeline}

\paragraph{Closed-Loop Execution.}
We formulate the single-step decision-making process of the embodied agent as a POMDP (see Appendix~\ref{app:pomdp_definition} for the complete tuple definition). The true environment state $s_t \in \mathcal{S}$ is not exposed to the policy; instead, the environment emits a partial observation $o_t^{\mathrm{env}}$ through the observation kernel $\Omega$. The agent operates in a continuous perception--reasoning--action loop. Given a task instruction $T$, the VLM policy produces a formatted tool call $a_t \in \mathcal{A}$ using tools free of oracle perceptual APIs. A tool executor validates and dispatches the call, immediately returning invalid actions as environmental feedback. This architecture decouples high-level decision-making from low-level safety verification and environmental dynamics.

\paragraph{Observation and State Aggregation.}
At decision step $t$, the environment first emits
\begin{equation}
o_t^{\mathrm{env}} \sim \Omega(\,\cdot \mid s_t).
\label{eq:env_observation}
\end{equation}
This environmental observation contains ego-centric visual inputs and execution feedback $e_t$ for the preceding action $a_{t-1}$. To avoid replaying the raw interaction trace, the policy combines it with a one-step action record and a compact long-term memory. Its complete conditioning context is
\begin{equation}
x_t =
\left(
T,\operatorname{Serialize}(\mathcal{A}),
o_t^{\mathrm{env}},a_{t-1},h_{t-1}
\right).
\label{eq:policy_context}
\end{equation}
We index the first decision by $t=1$ and initialize $a_0=\varnothing$, $\sigma_0=\varnothing$, $q_0=\operatorname{Init}(T)$, and $h_0=(\sigma_0,q_0)$. Thus, every component of $x_t$ is available before selecting $a_t$.

The model output contains a reasoning trace $c_t$, a compact \texttt{History} summary $\sigma_t$ of tool-verified action outcomes up to the current decision, an active task-phase analysis $q_t$, and the next action $a_t$. Denoting the model's assistant-response distribution by $\pi_\theta$, we write
\begin{equation}
(c_t,\sigma_t,q_t,a_t)
\sim \pi_\theta(\,\cdot \mid x_t),
\qquad
h_t=(\sigma_t,q_t).
\label{eq:policy_output}
\end{equation}
Unlike an unbounded concatenation of per-step messages, $\sigma_t$ is a single updated summary field. It may record both successful and failed actions, but only when their outcomes are grounded in the tool feedback $e_t$.

\paragraph{Scalable Asset Library and Task Instantiation.}
\label{sec:asset_library}
Our data engine leverages 7 high-fidelity scenes from GRScenes~\cite{wang2024grutopia} (11 room types, 100 interactive receptacles) and over 2,500 object instances spanning 639 categories from MesaTask~\cite{hao2025mesatask}. We represent each scene state as a \textbf{World Graph (WG)}, formally $\text{WG} = \{(r_j, \mathcal{E}_j)\}$, mapping each receptacle $r_j$ to its contained object instances $\mathcal{E}_j$. The WG serves as the canonical state representation throughout the framework: it grounds task definitions, enables state-differential reward computation (\cref{sec:reward_design}), and provides the success criterion for evaluation.

Tasks center on cross-receptacle object rearrangement: we prompt LLMs to generate semantically plausible \textit{(source, destination, object)} triplets conditioned on the scene layout, which are instantiated as task configurations by specifying an initial WG and a goal WG in the simulator.

To systematically challenge the agent, we introduce distractors at both the object and receptacle levels. Object-level distractors inject visually distinct instances of the same category to require fine-grained discrimination. At the receptacle level, each receptacle is annotated with \textit{uniquely identifying descriptions} covering functional purpose and visual appearance. LLMs then generate unambiguous natural-language instructions from these annotations. Detailed task-generation statistics and instruction examples are provided in Appendix~\ref{sec:supp_triplets} and Appendix~\ref{sec:supp_example_tasks}, respectively.

\paragraph{SFT Data Synthesis.}
For each task $i$, a rule-based planner decomposes the objective into an ordered action sequence $(a_t^{(i)})_{t=1}^{L_i}$ with $a_t^{(i)} \in \mathcal{A}$, covering navigation, exploration, and manipulation, and executes it in the simulator. At each step, the observation kernel $\Omega$ yields $o_t^{\mathrm{env}}$, including ego-centric RGB images with visual IDs and execution feedback $e_t$; all manipulation operates exclusively on \emph{visual IDs} from 2D perception.
In an automated \textbf{Social Annotation} phase for collaborative trajectories, LLMs rewrite instructions to be deliberately vague, injecting communicative actions (\texttt{ask}) with simulated user responses into trajectories to train the agent to communicate actively under uncertainty. 
The resulting SFT dataset is $\mathcal{D}=\{(T^{(i)}, \tau^{(i)})\}_{i=1}^{N}$, where $N$ is the number of expert trajectories and $T^{(i)}$ is the task instruction for trajectory $i$. The $i$-th trajectory is
\begin{equation}
\tau^{(i)}
=
\left(
    \big(o_t^{\mathrm{env},(i)},a_{t-1}^{(i)},h_{t-1}^{(i)},
c_t^{(i)},h_t^{(i)},a_t^{(i)}\big)
\right)_{t=1}^{L_i},
\label{eq:sft_trajectory}
\end{equation}
where $a_{t-1}^{(i)}$ explicitly records the previous action, while $h_{t-1}^{(i)}$ provides the long-term memory available before the current decision. Each trajectory contains $L_i$ step-wise decision examples: $c_t^{(i)}$ is the supervised reasoning trace, $h_t^{(i)}$ is the updated long-term historical state, and $a_t^{(i)}$ is the expert action executed next. The observations and actions come from planner rollouts in the simulator, while the corresponding reasoning traces and historical states are annotated by \texttt{gemini-3-pro}~\cite{gemini2025}. Thus, each trajectory of length $L_i$ yields $L_i$ multimodal prompt--response pairs for SFT. The annotation procedure is detailed in Appendix~\ref{sec:supp_expert_tra_ann}.

\subsection{Paradigm Alignment via SFT}
\label{sec:sft}

\paragraph{Unified Input-Output Paradigm.}
We select Qwen3-VL-8B-Instruct as the backbone VLM to support deployment on computationally constrained robotic systems, implementing a unified input-output structure for both training and inference. 

At each time step, the model receives the conditioning context $x_t$ defined in \cref{eq:policy_context}. The action-space schemas are serialized into the prompt, $o_t^{\mathrm{env}}$ provides the current visual observation and execution feedback, $a_{t-1}$ records the immediately preceding action, and $h_{t-1}$ supplies compact long-term memory. Given $x_t$, the VLM generates an explicit reasoning trace $c_t$ (e.g., \texttt{<think>...</think>}) followed by a JSON object containing the updated memory $h_t$ and next tool call $a_t\in\mathcal{A}$. We denote the tokenized serialization of this complete assistant response by
\begin{equation}
    y_t=\operatorname{Serialize}(c_t,h_t,a_t).
\label{eq:serialized_response}
\end{equation}
The same input--output format is used during training and inference: the annotated $h_t$ is included in the next step's historical input during teacher-forced SFT, whereas the model-generated $h_t$ is used during closed-loop deployment.

\paragraph{Reasoning and Tool-Calling Alignment via Behavior Cloning.}
In the first stage, we adapt the pre-trained VLM to the structured reasoning and tool interface by behavior cloning on $\mathcal{D}$. Let $\pi_\theta$ denote the autoregressive next-token distribution parameterized by the VLM. We minimize the token-level negative log-likelihood of the complete expert responses:
\begin{equation}
    \mathcal{L}_{\text{SFT}}
    =
    -\frac{1}{N_{\mathrm{tok}}}
    \sum_{i=1}^{N}\sum_{t=1}^{L_i}\sum_{k=1}^{|y_t^{(i)}|}
    \log \pi_\theta\!\left(
        y_{t,k}^{(i)}
        \mid x_t^{(i)},y_{t,<k}^{(i)}
    \right),
\end{equation}
where $y_{t,k}^{(i)}$ is the $k$-th token of the serialized response, $y_{t,<k}^{(i)}$ is its ground-truth prefix under teacher forcing, and $|y_t^{(i)}|$ is its token length. The normalization term $N_{\mathrm{tok}}=\sum_{i=1}^{N}\sum_{t=1}^{L_i}|y_t^{(i)}|$ is the total number of supervised response tokens. The summation covers all assistant-response tokens, including the reasoning trace, historical state, and tool call. The textual prompt and visual observation remain conditioning inputs but are excluded from the prediction targets. We freeze the vision encoder to preserve pre-trained visual representations.

\subsection{Reinforcement Learning for Skill Optimization}
\label{sec:rl}

Deploying the SFT-aligned initial policy $\pi_{\text{init}}$ (\cref{sec:sft}) in closed-loop inference exposes two limitations: (1)~\textit{distribution shift}, where behavior cloning on expert demonstrations does not equip the model to replan or recover from execution failures (\eg occluded targets, invalid API calls); and (2)~\textit{scarce interactive data}, where static datasets cannot cover the diversity of proactive human-robot interaction, causing the policy to overfit to templates rather than querying the user under ambiguity. RL addresses both issues by enhancing the agent to explore dynamic environments and actively acquire information from users.

\paragraph{Algorithm Selection.}
We aim to maximize the expected discounted return over variable-length trajectories under the POMDP formulation (\cref{sec:data_pipeline}). Accurately modeling a critic for VLM-based RL is computationally prohibitive, making critic-free methods such as Group Relative Policy Optimization (GRPO) a natural choice. However, GRPO applies importance ratios and clipping at the token level despite assigning sequence-level rewards. We therefore adopt GSPO, which constructs a length-normalized sequence ratio as the geometric mean of token-level ratios and applies clipping at the sequence level. This better aligns the optimization unit with the rollout reward and reduces sensitivity to token-level outliers. The objective and implementation details are provided in Appendix~\ref{app:training}.

\paragraph{Environment-Grounded Reward Shaping.}
\label{sec:reward_design}
Instead of sparse task-completion signals, we design a transition reward $r_t=\mathcal{R}(s_t,a_t,s_{t+1})$ grounded in the WG (\cref{sec:asset_library}). We define the goal-change set $\Delta_{\mathrm{goal}}=\text{WG}_{\mathrm{goal}}\ominus\text{WG}_{\mathrm{init}}$, where $\ominus$ enumerates signed additions, removals, and attribute changes (\eg picking and placing $n$ objects yields $2n$ entries). For reward shaping, we augment it with navigation and articulation targets $\Delta_{\mathrm{aux}}$ to obtain $\Delta_{\mathrm{task}}=\Delta_{\mathrm{goal}}\cup\Delta_{\mathrm{aux}}$. The observed event set $\delta_t$ for transition $(s_t,a_t,s_{t+1})$ is matched against the remaining targets in $\Delta_{\mathrm{task}}$; each new match yields positive reward and is marked complete to prevent duplicate credit. The physics engine inherently enforces temporal causality (\eg pick-before-place), eliminating the need to manually model temporal dependencies.

Beyond task progress, the agent receives a timing penalty $-\lambda_{\mathrm{time}}$, with $\lambda_{\mathrm{time}}>0$, and a severity-scaled penalty for failed actions based on the resulting traceback in $o_{t+1}^{\mathrm{env}}$. For tasks involving distractors, we reward the communicative tool \texttt{ask} subject to a difficulty-conditioned query budget $B_{\mathrm{ask}}\leq 3$. Detailed numerical configurations are provided in Appendix~\ref{app:reward}.

\section{Experiments}
\label{sec:experiments}

\subsection{Benchmark Construction}
To evaluate embodied agents along two key dimensions, \emph{active exploration} and \emph{communicative interaction}, we introduce \realbench. Built on the environment and tool interface (\cref{sec:environment}) and the data generation pipeline (\cref{sec:data_pipeline}), it covers novel scenarios across \real rooms. \realbench contains \textbf{241} task instances in four families: Furniture-Distractor Pick-and-Place (FDP, 72), Furniture \& Object-Distractor Pick-and-Place (FODP, 56), Furniture-Distractor Open/Close (FDO, 48), and Simulator-User-Loop (SUL, 65).

FDP and FODP are cluttered mobile manipulation tasks that require moving a target object from a source to a destination receptacle. FODP further adopts an open-vocabulary, full-category distractor setting at both the furniture and object levels, placing greater demands on active exploration and fine-grained visual disambiguation. FDO evaluates articulation-centric manipulation with openable and closeable furniture. These tasks involve highly dynamic state transitions and challenge temporal consistency and precise sequential tool use (\eg interleaving \texttt{open}/\texttt{close} with perception and pick-and-place actions). Finally, SUL introduces instruction ambiguity by design to test whether the agent can proactively query the simulated user via \texttt{ask} before execution.

\subsection{Results and Analysis of SFT and RL Training}

We evaluate (i) strong open-source and commercial VLMs under zero-shot prompting, and (ii) the proposed Qwen3-VL-8B agent after \emph{tool-interface alignment} via SFT and \emph{closed-loop} improvement via GSPO. \Cref{tab:main_results} reports \emph{success rates} (higher is better) on the four \realbench task families.

\begin{table*}[!htbp]
\centering
\small
\setlength{\tabcolsep}{5pt}
\renewcommand{\arraystretch}{1}
\caption{\textbf{Main results on \realbench.} Success rates (successful\,/\,total episodes) are reported across four task families. \textbf{Bold} marks the best result per column within each group, and the \colorbox{TeacherYellow}{highlighted} row denotes the teacher model used for SFT annotation. SFT aligns Qwen3-VL-8B to the tool interface free of oracle perceptual APIs, while GSPO further improves the interaction-heavy SUL split and surpasses all zero-shot VLM baselines, including the teacher.}
\label{tab:main_results}
\begin{tabularx}{\linewidth}{>{\raggedright\arraybackslash}p{0.43\linewidth} *{4}{>{\centering\arraybackslash}X}}
\toprule
\textbf{Model Name} & \textbf{FDP} & \textbf{FODP} & \textbf{FDO} & \textbf{SUL} \\
\midrule
\multicolumn{5}{c}{\textbf{Zero-shot Prompting}} \\
\midrule
\rowcolor{TeacherYellow}
gemini-3-pro-preview~\cite{gemini3pro}         & \textbf{81.9} & 39.3          & 50.0          & 53.8 \\
gemini-2.5-pro~\cite{gemini25report}           & 66.7          & \textbf{41.1} & 43.8          & 49.2 \\
gpt-5~\cite{openai2025gpt5}                    & 68.1          & 30.4          & 47.9          & 52.3 \\
gpt-4o~\cite{openai2024gpt4o}                  & 30.6          & 28.6          & 27.1          & 40.0 \\
claude-haiku-4-5~\cite{anthropic2025haiku45}   & 45.8          & 16.1          & \textbf{52.1} & 47.7 \\
Qwen3-VL-235B-A22B-Instruct~\cite{qwen3vl2025} & 22.2          & 14.3          & 14.6          & 18.5 \\
Qwen3-VL-8B-Instruct~\cite{qwen3vl2025}        & 0.0           & 0.0           & 0.0           & 1.5 \\
\midrule
\multicolumn{5}{c}{\textbf{Ours}} \\
\midrule
\textbf{Ours-8B-SFT-only (1 epoch)}   & 45.8 & 30.4 & 35.4 & 36.9 \\
\textbf{Ours-8B-SFT-only (2 epochs)}  & 65.3 & 28.6 & 43.8 & 50.8 \\
\textbf{Ours-8B-SFT+RL}               & 58.3 & 33.9 & 41.7 & \textbf{56.9} \\
\bottomrule
\end{tabularx}
\end{table*}

\paragraph{Tool Alignment and Behavior-Cloning Overfitting.}
Zero-shot Qwen3-VL-8B fails on manipulation tasks and achieves only 1.5\% on interactive scenarios. One epoch of supervised fine-tuning aligns the model with the tool interface, reaching 45.8\% on FDP. However, a second epoch exposes the limits of behavior cloning: while execution stability on standard manipulation improves (FDP rises to 65.3\%), performance on the distractor-rich FODP split drops from 30.4\% to 28.6\%. This suggests that strict imitation overfits expert trajectories and weakens robustness under open-vocabulary clutter and visual ambiguity.

\paragraph{Reinforcement Learning Restores Generalization.}
Applying closed-loop group sequence policy optimization to the single-epoch checkpoint circumvents this overfitting. RL recovers the exploration capability lost during extended supervised training, increasing FODP success to 33.9\% while maintaining tool-calling performance comparable to the model more tightly fine-tuned on standard tasks. This suggests that environmental feedback promotes adaptive problem-solving over rote trajectory memorization, helping the agent handle edge cases.

\paragraph{Social Reward Improves Proactive Intent Alignment.}
The impact of closed-loop optimization is most pronounced in the interaction-heavy SUL split, where the agent reaches 56.9\% success and outperforms the strongest proprietary zero-shot baselines. By integrating a difficulty-conditioned query budget and explicit rewards for resolving ambiguity, the training pipeline shifts the agent from passive instruction following to proactive intent alignment. The policy learns when to pause and query the simulator-user, showing that social interaction emerges from end-to-end environmental feedback rather than static prompt engineering.

\begin{wrapfigure}{r}{0.48\linewidth}
  \centering
  \includegraphics[width=\linewidth,trim=200 0 10 5,clip]{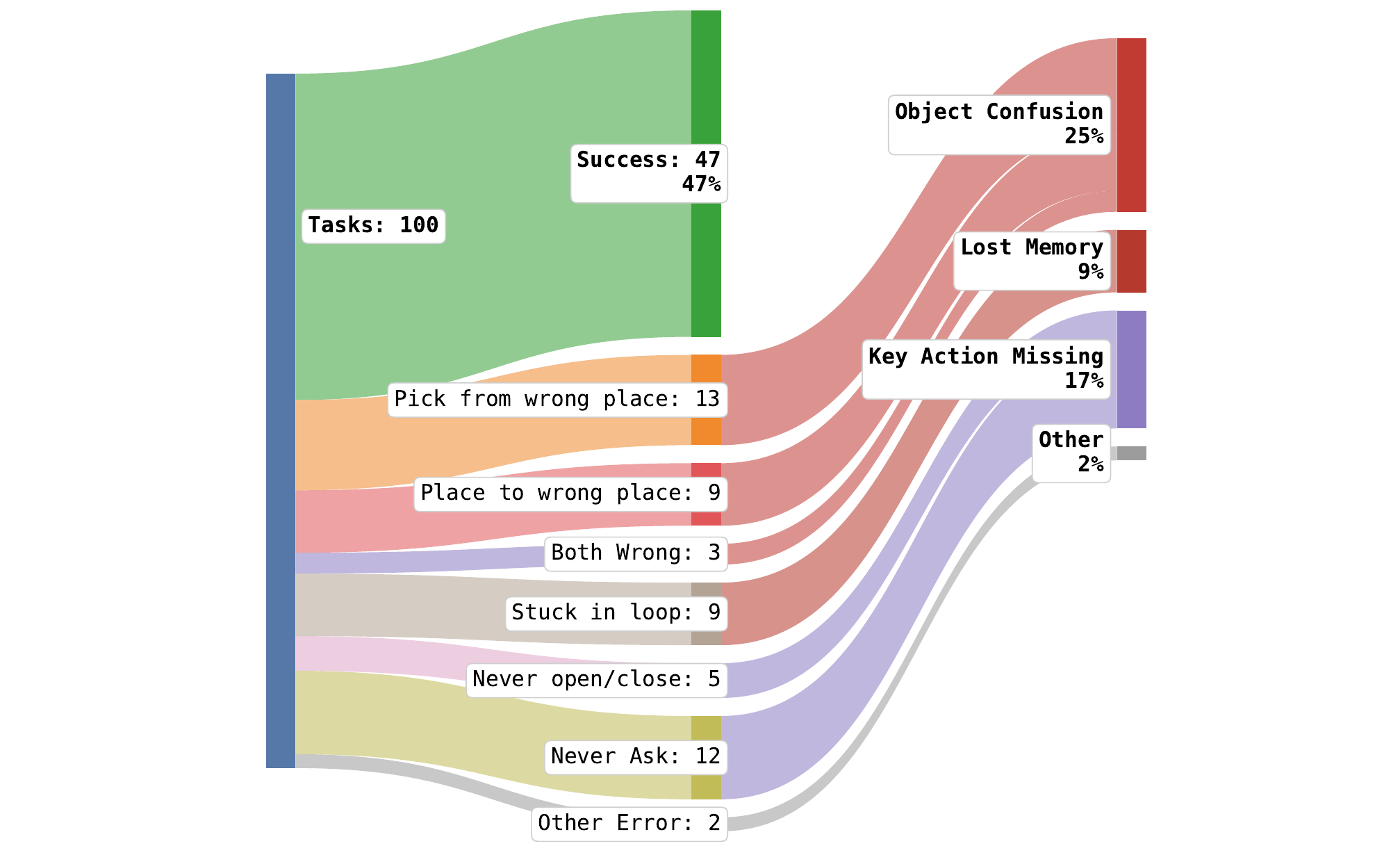}
  \caption{\textbf{Failure mode taxonomy.} Manual analysis over 100 evaluation episodes from the best SFT+RL checkpoint.}
  \label{fig:failure_sankey}
\end{wrapfigure}

\raggedright
\paragraph{Complementary Effects of SFT and RL Reveal a Persistent Vision Bottleneck.}
SFT substantially improves performance across all four visually grounded task families over the base model, indicating better adaptation to embodied visual observations and the tool-use interface. However, the decrease on FODP from 30.4\% after one epoch to 28.6\% after two epochs suggests a risk of behavior-cloning overfitting. Starting from the one-epoch SFT checkpoint, RL improves every split, with the largest gain on the interaction-heavy SUL benchmark (36.9\% to 56.9\%), while also improving exploration and multi-step task execution. Nevertheless, because the vision encoder remains frozen, RL can only influence visual grounding indirectly and cannot directly refine the underlying visual representations. The dominance of object-confusion errors therefore indicates that fine-grained visual grounding remains a key bottleneck, motivating more explicit perceptual supervision or selective adaptation of the vision module.\par
\justifying

\paragraph{Failure Mode Taxonomy.}
We manually annotate 100 episodes from the best SFT+RL checkpoint to diagnose residual errors (\cref{fig:failure_sankey}). Of 53 failures, most fall under \emph{Object Confusion} (25), followed by \emph{Key Action Missing} (17), \emph{Lost Memory} (9), and \emph{Other Error} (2). These results identify cluttered visual grounding, procedural completeness, and long-horizon memory as the main remaining bottlenecks.

\FloatBarrier
\subsection{Real-World Deployment via Sim-to-Real Transfer}
\label{sec:real_world}

\begin{figure*}[t]
  \centering
  \includegraphics[width=\linewidth]{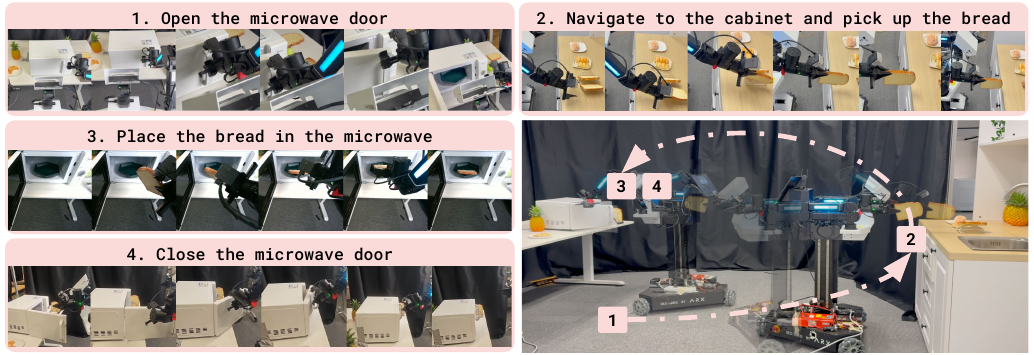}
  \caption{\textbf{Real-world deployment on the ARX LIFT2 mobile manipulator.} \real transfers the high-level visual-interactive policy zero-shot to a physical backend with the same MCP tool interface. The execution trajectory demonstrates causal planning (\eg preparing the microwave before fetching food), human-robot disambiguation under multiple candidate objects, and language-to-physical grounding for closed-loop task completion. Quantitative reliability is reported in \cref{tab:real_world_compact}.}
  \label{fig:real_world}
\end{figure*}

To validate the effectiveness and robustness of our training pipeline, we deploy the simulated agent in a real-world household scenario using the ARX LIFT2 mobile manipulator. This direct sim-to-real transfer tests how capabilities learned in simulation map to physical execution across varying environmental complexity.

\paragraph{Seamless Transfer Through Standardized Interfaces.}
The system uses a Model Context Protocol(MCP) server for tool invocation, keeping the high-level cognitive framework identical between simulation and the physical robot. At each step, agent receives real-time ego-centric observations, combines them with the structured history and immediate past action, and outputs the next discrete tool call. This standardized interface decouples the reasoning algorithm from the environment, allowing the physical world to act as an alternative backend to the simulator without retraining the high-level policy.

\paragraph{Hierarchical Execution with Specialized Physical Policies.}
While high-level planning is transferred zero-shot from simulation, continuous physical dynamics still require robust motor control. We therefore map the abstract tools used in simulation to dedicated real-world implementations. Specifically, atomic manipulation commands are executed by a fine-tuned $\pi_{0.5}$~\cite{intelligence2025pi05} base model. This hierarchy keeps VLM focused on visual grounding, state tracking, and task planning, while the $\pi_{0.5}$ policy provides stable execution for opening, closing, picking, and placing.

\paragraph{Long-Horizon Causal Reasoning in Physical Environments.}
As shown in \cref{fig:real_world}, the system successfully executes a multi-stage food preparation task requiring the robot to fetch bread and place it in a microwave. Rather than immediately navigating to the cabinet, the reasoning engine first approaches the table and opens the microwave door. Only after preparing the receptacle does the robot navigate to the secondary location, select the target object, and return to complete placement and closure. This sequence suggests that the structured memory mechanism helps prevent temporal hallucinations, maintain an accurate internal state, and respect causal constraints over long physical executions.

\paragraph{Physical Evaluation and Pipeline Analysis.}
We further evaluate the pipeline in the physical environment across different spatial layouts. Across 60 real-world episodes, \real achieves 78.3\% end-to-end success with zero unrecoverable system crashes (\cref{tab:real_world_compact}). The low-level primitive layer is executable in 85.3\% of 600 VLA primitive executions, indicating that the high-level policy transfers without requiring privileged simulator state while still depending on robust physical skills. A more detailed breakdown of these real-world trials, along with primitive-level evaluation and failure analysis, is provided in Appendix~\ref{subsec:supp_real_world_quant}.

\begin{table}[t]
\centering
% \scriptsize
% \setlength{\tabcolsep}{4pt}
% \renewcommand{\arraystretch}{0.94}
\caption{\textbf{End-to-end real-world evaluation.} SR denotes task success rate, SPL measures path efficiency, Ask is the average number of user queries, and VLM Lat. reports total VLM inference latency per episode.}
\label{tab:real_world_compact}
\begin{tabular}{lccccc}
\toprule
Task & SR & Steps & SPL & Ask & VLM Lat. \\
\midrule
FDO & 80.0\% & 12.2 & 67.0\% & 0.90 & 71.1s \\
FODP & 100.0\% & 9.6 & 86.9\% & 0.70 & 53.5s \\
SUL & 55.0\% & 13.0 & 35.4\% & 0.55 & 80.5s \\
\midrule
\textbf{Overall} & \textbf{78.3\%} & \textbf{11.6} & \textbf{63.1\%} & \textbf{0.72} & \textbf{68.4s} \\
\bottomrule
\end{tabular}
\end{table}

\section{Conclusion and Limitations}
\label{sec:conclusion}

We presented \real, a closed-loop framework for training exploratory and communicative embodied agents that transfer directly to physical robots. By replacing oracle perceptual APIs with a physically realizable exploration toolchain and integrating an VLM-driven simulated user for dynamic intent alignment, \real addresses two key oversimplifications in prior work. Training a Qwen3-VL-8B agent via SFT followed by closed-loop GSPO on \realbench (241 tasks, four families) shows that tool-interface alignment is a prerequisite for closing the perception-action loop, while RL uniquely improves interactive disambiguation. Deployment on the ARX LIFT2 dual-arm mobile robot further validates sim-to-real transferability for long-horizon household tasks.

Despite these results, several limitations remain.
First, the task space centers on cross-receptacle rearrangement and does not yet encode richer compositional constraints such as temporally ordered sub-tasks~\cite{chang2024partnr} or spatial relational goals~\cite{dai2025manitaskgen} (\eg ``place the cup \emph{to the left of} the plate'').
Second, the simulated user operates within a bounded behavioral envelope---its preferences are constrained to in-scene objects and its responses are anchored to the agent's trajectory---whereas real users may introduce new goals mid-task or convey intent through implicit cues.
Third, receptacles are modeled as monolithic entities (\eg \texttt{cabinet\_1}) without part-level distinction (\eg individual shelves), limiting spatial precision for both exploration and manipulation. Incorporating part-level segmentation and hierarchical receptacle modeling would address this gap.

\section{Acknowledgments}
\label{sec:ac}
This work was supported in part by Shanghai Artificial Intelligence Laboratory, the National Natural Science Foundation of China under Grant 62401367 and the Shanghai Magnolia Talent Program Pujiang Project under Grant No. 25PJA076.

\clearpage
\appendix
\section*{Appendix}
\addcontentsline{toc}{section}{Appendix}
\suppressfloats[t]
\setcounter{section}{0}
\renewcommand{\thesection}{\Alph{section}}
\renewcommand{\thesubsection}{\thesection.\arabic{subsection}}
\renewcommand{\thesubsubsection}{\thesubsection.\arabic{subsubsection}}
\renewcommand{\thefigure}{\thesection.\arabic{figure}}
\renewcommand{\thetable}{\thesection.\arabic{table}}
\renewcommand{\thealgorithm}{\thesection.\arabic{algorithm}}
\newcommand{\inputappendixsection}[1]{%
  \FloatBarrier
  \setcounter{figure}{0}%
  \setcounter{table}{0}%
  \setcounter{algorithm}{0}%
  \input{#1}%
}
\FloatBarrier

  \FloatBarrier
  \setcounter{figure}{0}%
  \setcounter{table}{0}%
  \setcounter{algorithm}{0}%
  \section{Scene Setup and Task Synthesis}
\label{sec:scenes}

\subsection{Training Scenes}

\begin{figure}[ht]
    \centering
    \includegraphics[width=0.45\linewidth]{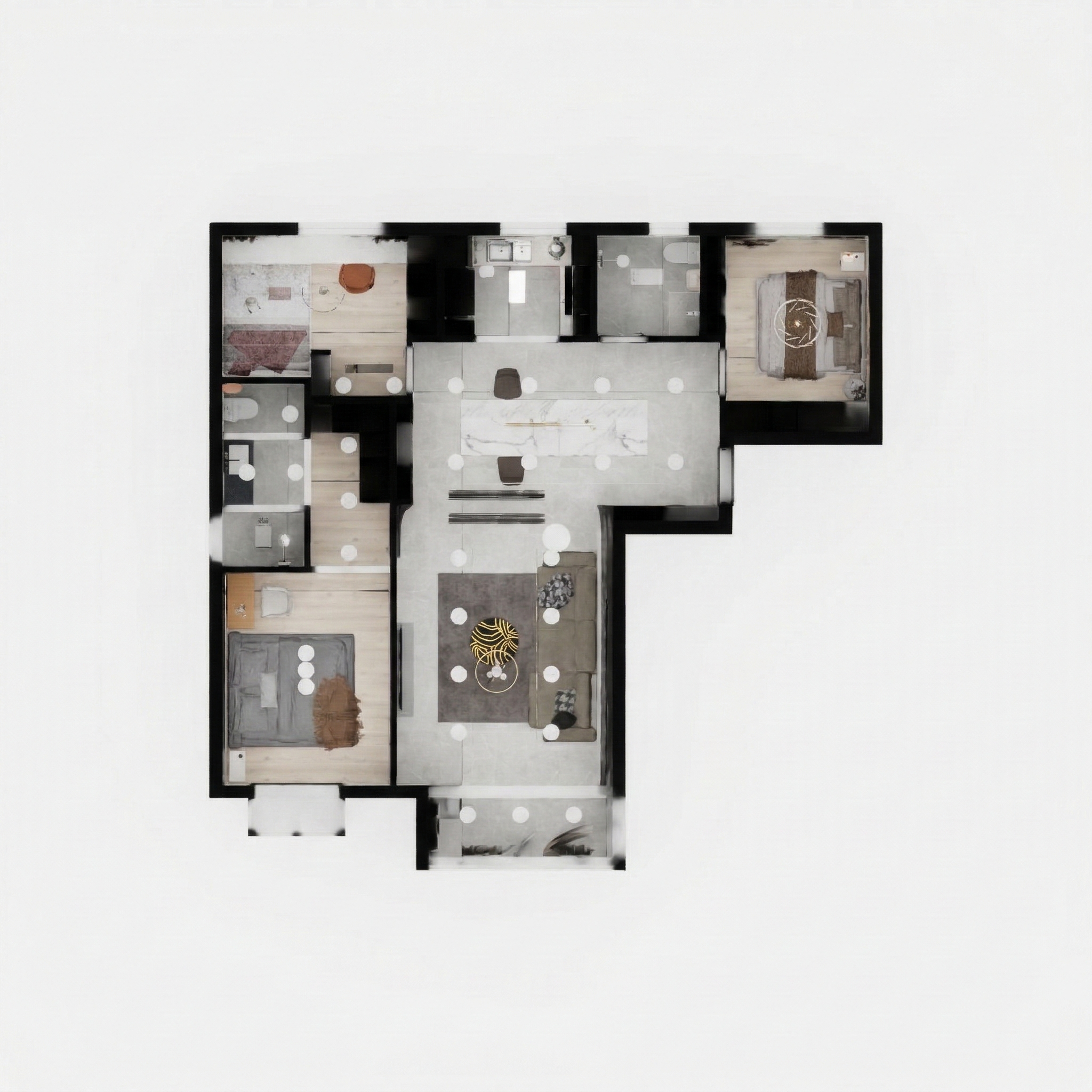}
    \hfill
    \includegraphics[width=0.45\linewidth]{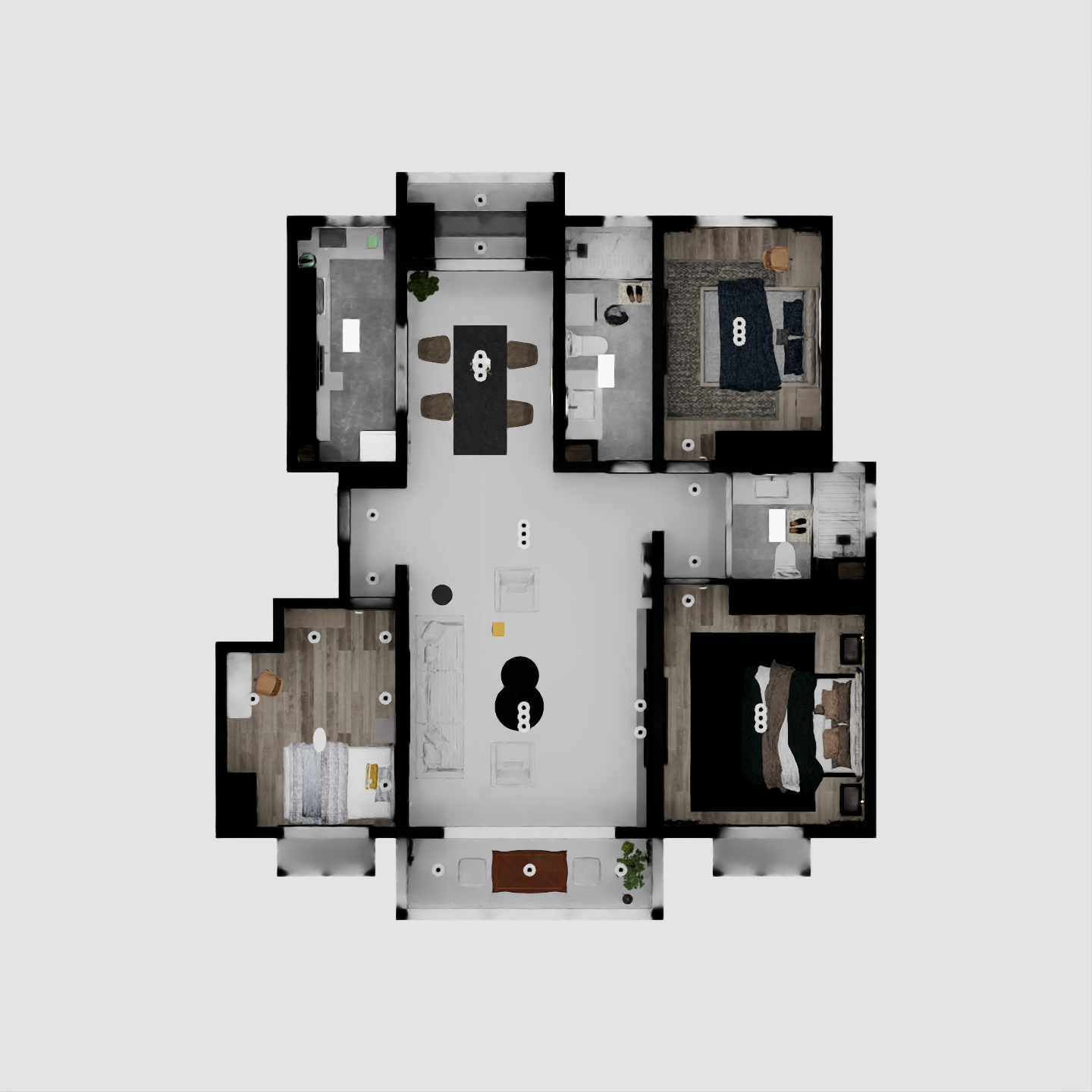}
    \\
    \caption{
    \textbf{Training scenes:} Overhead floor plan views of two training environments. The scenes display the spatial layout along with the arrangements of furniture and objects. They vary in complexity, room configurations, and object density, offering diverse training examples to learn capabilities in spatial reasoning, active perception, and manipulation.
    }
    \label{fig:training_scenes}
\end{figure}

The training and evaluation datasets are constructed using high-fidelity 3D scenes sourced from GRScene~\cite{wang2024grutopia}.
\Cref{fig:training_scenes} presents overhead floor plan views of two training environments, illustrating the spatial layout and object arrangements.
The training set comprises \textbf{6 residential indoor environments}, featuring a total of \textbf{82 furniture pieces} and \textbf{157 natural-language captions}.
The complexity of the scenes ranges from 5 to 7 room types and 10 to 17 furniture pieces per scene. Each scene contains multiple rooms with various furniture pieces, offering diverse spatial configurations that facilitate the learning of mobile manipulation and spatial reasoning.
\Cref{tab:training_scenes} details the breakdown of rooms, furniture, and captions for each scene.

\begin{table}[H]
    \centering
    \caption{Per-scene statistics for the 6 training environments. \#Furniture denotes the number of interactable receptacle pieces; \#Captions denotes the total number of annotated natural-language descriptions.}
    \label{tab:training_scenes}
    \setlength{\tabcolsep}{3pt}
    \small
    \begin{tabular}{@{}cp{3.6cm}p{3.8cm}cc@{}}
        \toprule
        \textbf{Scene} & \textbf{Room Types} & \textbf{Furniture Types} & \textbf{\#Furn.} & \textbf{\#Cap.} \\
        \midrule
        S1 & balcony, bedroom, dining room, kitchen, living room, study room, toilet
           & bed, cabinet, chest of drawers, desk, microwave, nightstand, refrigerator, shoe cabinet, table, tea table
           & 17 & 32 \\
        \midrule
        S2 & balcony, bedroom, dining room, entry, kitchen, living room
           & bed, cabinet, couch, desk, fridge, microwave, table
           & 17 & 29 \\
        \midrule
        S3 & balcony, bedroom, dining room, kitchen, living room
           & bed, couch, desk, microwave, refrigerator, table, TV stand
           & 12 & 21 \\
        \midrule
        S4 & bathroom, bedroom, corridor, dining room, kitchen, living room
           & bed, cabinet, couch, desk, microwave, refrigerator, table
           & 12 & 24 \\
        \midrule
        S5 & bedroom, dining room, kitchen, living room, toilet
           & bed, cabinet, couch, counter, desk, microwave, refrigerator, table
           & 14 & 32 \\
        \midrule
        S6 & bedroom, dining room, kitchen, living room, toilet
           & bed, cabinet, refrigerator, sink, table, TV stand
           & 10 & 19 \\
        \midrule
        \textbf{Total} & & & \textbf{82} & \textbf{157} \\
        \bottomrule
    \end{tabular}
\end{table}

\subsection{Evaluation Scene}

The test set consists of a held-out scene (S7) used to evaluate the model's generalization to previously unseen spatial configurations. This scene is specifically selected to establish a meaningful evaluation benchmark while introducing a distribution shift from the training scenes. S7 contains 6 room types (bedroom, dining room, kitchen, living room, study room, and washroom) along with 18 furniture pieces. Furthermore, it incorporates articulated appliances (such as a washing machine and an electric cooker) that are not prominently featured during training.

\begin{figure}[ht]
    \centering
    \includegraphics[width=0.6\linewidth]{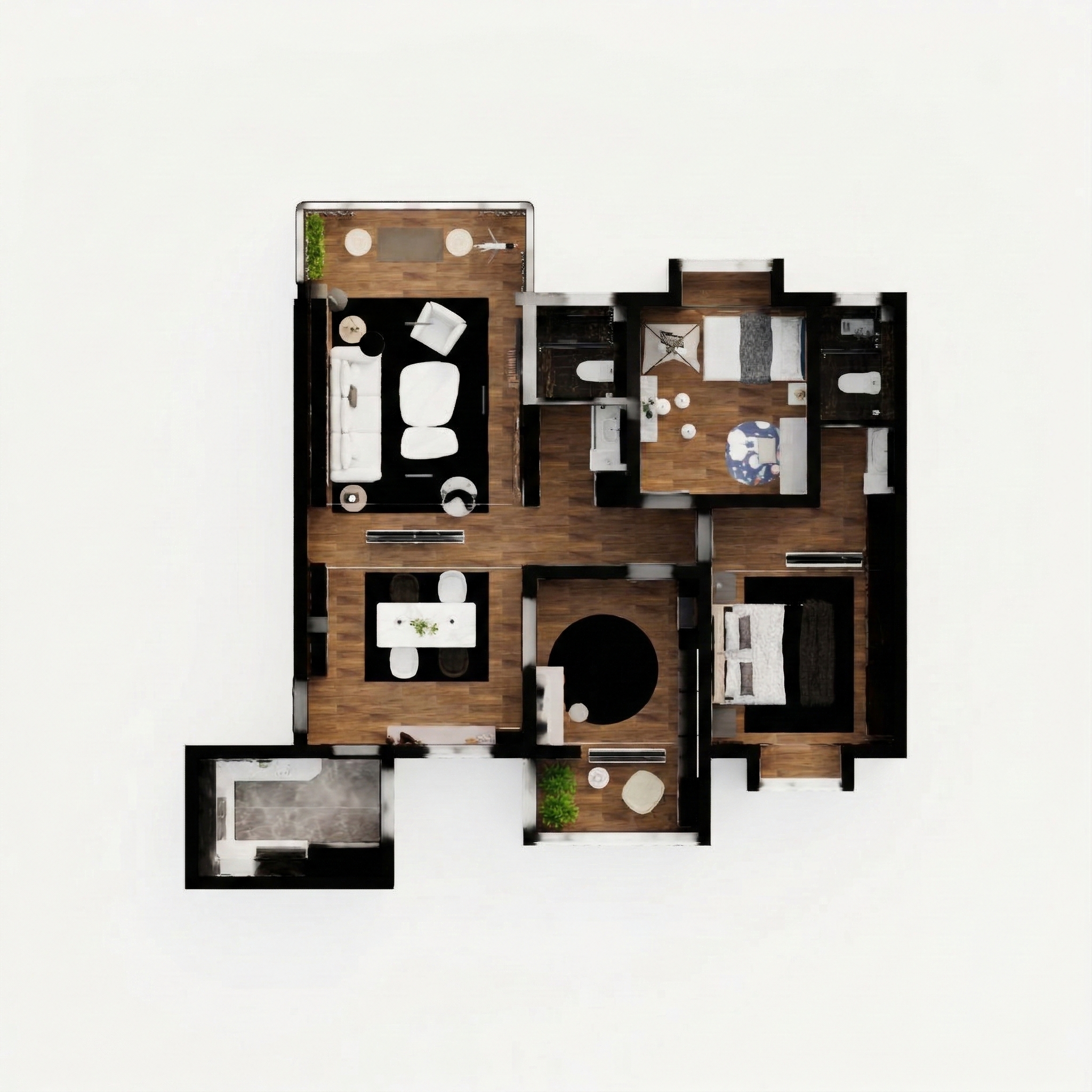}
    \\
    \caption{
    \textbf{Evaluation scene:} Overhead floor plan of the evaluation environment. This scene is held out from training and used to assess the performance of the model on unseen spatial layouts. The composition of the scene and the arrangements of objects present different spatial challenges compared to the training set, thereby enabling a fair evaluation of generalization.
    }
    \label{fig:test_scene}
\end{figure}

\subsection{3D Asset Library}
\label{sec:supp_asset_library}

To support the generation of diverse and realistic layouts, a large-scale asset library is curated, comprising manipulable objects from MesaTask~\cite{hao2025mesatask}.

\paragraph{Object Assets.}
The object asset library contains \textbf{3,204 unique 3D assets} that span \textbf{810 fine-grained semantic categories}. As summarized in Table~\ref{tab:asset_categories}, these assets are organized into four general categories: \textit{household items} (1,865 assets), \textit{containers} (1,185 assets), \textit{food} (127 assets), and \textit{furniture} (27 assets). 
The experiments primarily use the assets from the \textit{household items} and \textit{containers} categories.
Each asset is annotated with physical size information, semantic category labels, and detailed descriptions. This broad coverage ensures that the generated tasks reflect the diversity of everyday household manipulation scenarios, ranging from placing food items into containers to rearranging household objects across different room types.

\begin{table}[h]
    \centering
    \caption{Distribution of object assets across general categories.}
    \label{tab:asset_categories}
    \begin{tabular}{lcc}
        \toprule
        \textbf{General Category} & \textbf{Asset Count} & \textbf{Proportion} \\
        \midrule
        Household Items  & 1,865 & 58.2\% \\
        \midrule
        Containers       & 1,185 & 37.0\% \\
        \midrule
        Food             & 127   & 4.0\%  \\
        \midrule
        Furniture        & 27    & 0.8\%  \\
        \midrule
        \textbf{Total}   & \textbf{3,204} & \textbf{100\%} \\
        \bottomrule
    \end{tabular}
\end{table}

% \paragraph{Scene and Furniture Assets.}
% Indoor scenes are drawn from high-fidelity 3D residential environments. Our dataset includes \textbf{7 distinct scenes} covering \textbf{11 room types}: living room, bedroom, dining room, kitchen, study room, bathroom, balcony, corridor, entry, toilet, and washroom. Each scene is populated with interactable furniture serving as receptacle surfaces, including cabinets (21 instances), beds (19), tables (17), desks (13), refrigerators (5), microwaves (5), couches (5), sinks (3), TV stands (2), and various other household appliances. Each furniture piece is annotated with its room assignment, surface receptacle bounding box, and natural-language captions describing its appearance, enabling the generation of spatially grounded and linguistically diverse manipulation instructions.

\subsection{Semantic-Aware Task Generation}
\label{sec:supp_triplets}

As described in \cref{sec:asset_library}, the tasks are defined by LLM-generated \textit{(source, destination, object)} triplets. These triplets specify a source receptacle for picking, a destination receptacle for placing, and a set of semantically plausible objects for the transfer. The generation of triplets is conditioned on the layout of the scene and the inter-room semantic compatibility, which yields 291 unique receptacle-pair groups that cover diverse scenarios for room-to-room rearrangement. Table~\ref{tab:triplet_examples} presents a selection of examples that illustrate the spatial and semantic diversity of the generated tasks.

\begin{table}[!t]
    \centering
    \caption{
        Examples of \textit{(source, destination, object)} triplets used to instantiate tasks.
        Each row corresponds to a single receptacle pair; the \textit{Objects} column lists a sample of semantically compatible items for the specified transfer.
        The source and destination are denoted as \textit{room\,/\,furniture}.
    }
    \label{tab:triplet_examples}
    \setlength{\tabcolsep}{4pt}
    \begin{tabular}{>{\raggedright\arraybackslash}p{3.2cm}
                    >{\raggedright\arraybackslash}p{3.4cm}
                    >{\raggedright\arraybackslash}p{4.7cm}}
        \toprule
        \textbf{Source} & \textbf{Destination} & \textbf{Objects (sample)} \\
        \midrule
        kitchen / refrigerator    & dining room / table         & drink, apple, cake, juice \\
        \midrule
        dining room / table       & kitchen / refrigerator      & apple, butter, fruit, dessert \\
        \midrule
        bedroom / bed             & bedroom / nightstand        & alarm clock, book, notebook, remote control \\
        \midrule
        study room / desk         & bedroom / cabinet           & notebook, pen holder, picture frame, book \\
        \midrule
        living room / table       & living room / TV stand      & speaker, book, basket, remote control \\
        \midrule
        bathroom / sink           & bathroom / cabinet          & towel, napkin, tray, hand soap \\
        \midrule
        balcony / table           & living room / cabinet       & basket, flower, book, ornament \\
        % balcony / table           & bedroom / cabinet           & basket, pen, book, clock \\
        % bedroom / nightstand      & kitchen / dining table      & apple, cake, fruit, water glass \\
        % dining room / table       & washingroom / sink          & towel, napkin, tray, hand soap \\
        \bottomrule
    \end{tabular}
\end{table}

\subsection{Task Instruction Examples}
\label{sec:supp_example_tasks}

The tasks are grouped into two categories. \textbf{Static tasks} require no human interaction and target the core manipulation capabilities of the agent, including spatial reasoning, object recognition, and goal-directed navigation. \textbf{Dynamic tasks} require the agent to interact with a human user to disambiguate the true goal of the task, thereby targeting the communicative and clarification capabilities of the agent. Table~\ref{tab:example_tasks} provides illustrative examples. The color \textcolor{figblue}{blue} highlights the source or destination furniture, while \textcolor{figpink}{red} highlights the target object.

\begin{table}[!t]
    \centering
    \caption{
        Task examples grouped by interaction type.
        \textit{Precise} instructions reference objects with visual attributes, whereas \textit{fuzzy} instructions use only category labels.
        The dynamic tasks employ deliberately vague language to elicit clarification dialogues.
        \textcolor{figblue}{Blue}: source and destination furniture. \textcolor{figpink}{Red}: target object.
    }
    \label{tab:example_tasks}
    \setlength{\tabcolsep}{4pt}
    \small
    \begin{tabular}{p{2.4cm} p{8.6cm}}
        \toprule
        \textbf{Split} & \textbf{Task Instruction} \\
        \midrule

        \multirow{4}{*}{\parbox{2.2cm}{\textbf{Static Task}}}
        & \textit{``Head over to \textcolor{figblue}{bedside cabinet with thin handles}, fetch the \textcolor{figpink}{rectangular wooden picture frame with a clean design and light finish}, and bring it back to \textcolor{figblue}{display cabinet with circular decor}.''}\\[3pt]
        \cmidrule(l){2-2}
        & \textit{``Pick up the \textcolor{figpink}{sleek black remote control with a circular navigation button and multicolored buttons} from \textcolor{figblue}{gold leg coffee table} and place it on \textcolor{figblue}{desk with white marble tabletop}.''}\\[3pt]
        \cmidrule(l){2-2}
        & \textit{``Head over to \textcolor{figblue}{two drawer cabinet}, fetch an \textcolor{figpink}{avocado}, and bring it back to the only \textcolor{figblue}{refrigerator} in the scene.''}\\[3pt]
        \cmidrule(l){2-2}
        & \textit{``Head over to the only \textcolor{figblue}{refrigerator} in the scene, fetch a \textcolor{figpink}{peanut}, and bring it back to the only \textcolor{figblue}{tea table} in the room.''}\\

        \midrule

        \multirow{3}{*}{\parbox{2.2cm}{\textbf{Dynamic Task}}}
        & \textit{``Bring a \textcolor{figpink}{notebook} from \textcolor{figblue}{desk with ribbed drawer front} to \textcolor{figblue}{cabinet with dark interior backing}.''}\\[3pt]
        \cmidrule(l){2-2}
        & \textit{``Can you find my \textcolor{figpink}{pen holder} in the bedroom?''}\\[3pt]
        \cmidrule(l){2-2}
        & \textit{``Can you help me move a \textcolor{figpink}{book}?''}\\

        \bottomrule
    \end{tabular}
\end{table}
  \FloatBarrier
  \setcounter{figure}{0}%
  \setcounter{table}{0}%
  \setcounter{algorithm}{0}%
  % ============================================================
%  Appendix: Data Generation Pipeline Details
% ============================================================

\section{Training Data Generation Pipeline}
\label{sec:supp_pipeline}

This section describes the three-stage data generation pipeline that transforms raw task specifications into the final SFT training corpus: (1) two-stage geometric and physics-based task filtering, (2) simulation-driven oracle trajectory generation, and (3) LLM-based social interaction annotation.

% ------------------------------------------------------------
\subsection{Collision-Free Object Placement}
\label{sec:supp_placement}

\begin{figure}[h]
    \centering
    \includegraphics[width=0.6\linewidth]{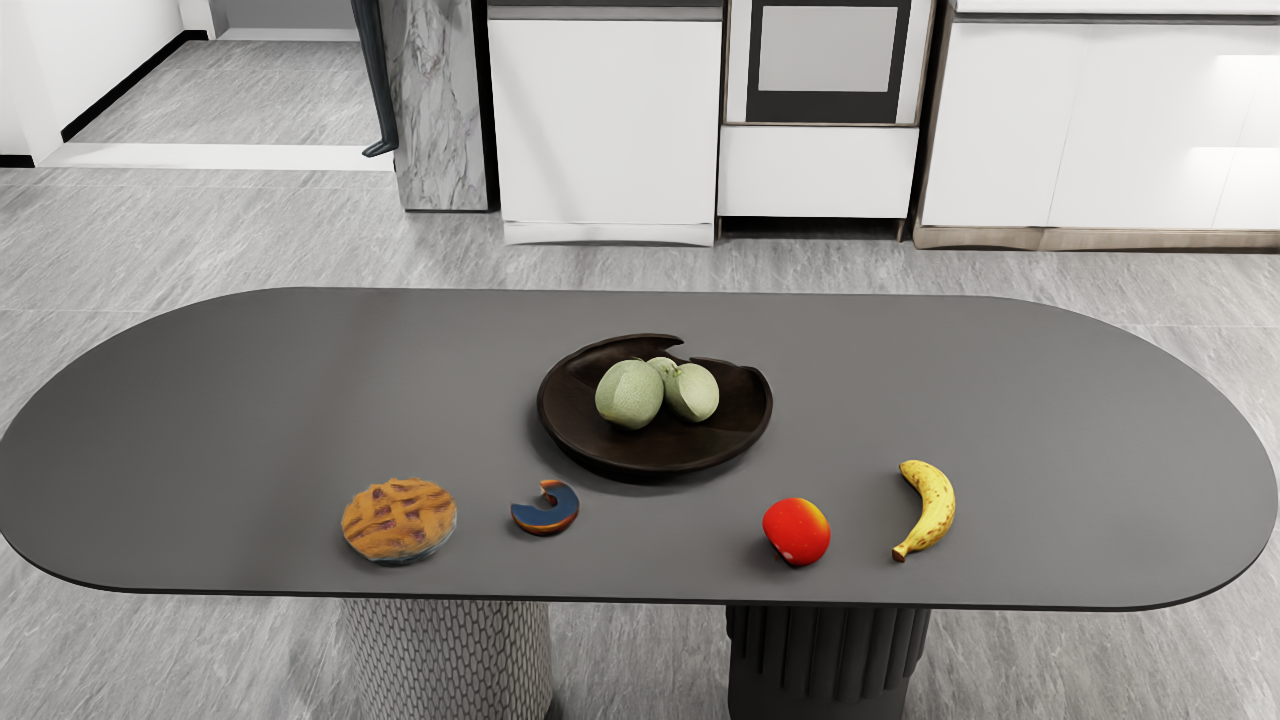}
    \caption{Example object layout produced by the placement algorithm on a kitchen table. Five food items are distributed across the surface, each within the perimeter reachability band and clear of static obstacles.}
    \label{fig:demo_layout}
\end{figure}

Before filtering, each task specification must be grounded into concrete 3D object positions.
Given a receptacle surface with bounding box $\mathcal{B}$ and the scene's pre-computed occupancy map $\mathcal{M}$ (a 2D grid encoding free vs.\ occupied cells at $\sim$5\,cm resolution), the placement algorithm assigns a collision-free 2D position to each object while satisfying two constraints beyond mere non-overlap:
(i) \emph{free-space feasibility}---the circular footprint of radius $r_i = \tfrac{1}{2}\sqrt{w_i^2 + d_i^2}$ centered at the chosen position must lie entirely within free cells of $\mathcal{M}$, ensuring no static obstacle occludes the object; and
(ii) \emph{edge-proximity for reachability}---each object is restricted to a perimeter band of width $d_{\max} = 0.5$\,m from the nearest receptacle edge, and the edge point closest to the object must also be unoccupied within a small robot-stand radius of 5\,cm, guaranteeing that a mobile manipulator can approach and grasp the object without obstruction.
Candidate positions are generated by uniform grid sampling over axis-aligned edge strips that tile the perimeter band; among all valid candidates per iteration, the algorithm selects the one maximizing the minimum Euclidean distance to already-placed objects, promoting spatial spread.
Figure~\ref{fig:demo_layout} shows a representative layout produced by this procedure.

% ------------------------------------------------------------
\subsection{Two-Stage Task Filtering}
\label{sec:supp_filtering}

Raw task specifications contain object placements that may be geometrically infeasible or physically unstable in the target 3D scene.
We apply a two-stage filtering pipeline to ensure every training episode is physically realizable and reproducible.

\paragraph{Stage~1: Static Geometric Filtering.}
The static filter checks whether all objects in a task's initial world graph can be placed on their assigned receptacle surfaces \emph{without} launching the physics simulator.
For each receptacle, we retrieve its surface bounding box $\mathcal{B} = [\mathbf{p}_{\min},\,\mathbf{p}_{\max}]$ from the scene furniture library and the physical footprint $(w_i, d_i, h_i)$ of each object $i$ from the asset size library.
A randomized collision-free placement algorithm (\cref{alg:static_filter}) then attempts to assign a 2D position to each object within $\mathcal{B}$, with a minimum inter-object margin of $\delta = 0.05$\,m.
 
\begin{algorithm}[H]
\caption{Static Placement Feasibility Check}
\label{alg:static_filter}
\small
\begin{algorithmic}[1]
\Require Object list $\mathcal{V}_{\mathrm{obj}}$, receptacle bbox $\mathcal{B}$, margin $\delta$, max attempts $K_{\mathrm{try}}$
\Ensure Placement dictionary $\mathcal{Q}$ or \textsc{Fail}
\State $\mathcal{Q} \leftarrow \{\}$;\quad $\text{placed} \leftarrow []$
\For{each object $o_i \in \mathcal{V}_{\mathrm{obj}}$ with footprint $(w_i, d_i)$}
    \State $x_{\text{lo}} \leftarrow p_{\min}^x + w_i/2 + \delta$;\quad $x_{\text{hi}} \leftarrow p_{\max}^x - w_i/2 - \delta$
    \State $y_{\text{lo}} \leftarrow p_{\min}^y + d_i/2 + \delta$;\quad $y_{\text{hi}} \leftarrow p_{\max}^y - d_i/2 - \delta$
    \If{$x_{\text{lo}} \geq x_{\text{hi}}$ \textbf{or} $y_{\text{lo}} \geq y_{\text{hi}}$}
        \Return \textsc{Fail} \Comment{Object too large for surface}
    \EndIf
    \State $\text{found} \leftarrow \text{False}$
    \For{$k = 1$ \textbf{to} $K_{\mathrm{try}}$}
        \State Sample $(x, y) \sim \mathcal{U}([x_{\text{lo}}, x_{\text{hi}}] \times [y_{\text{lo}}, y_{\text{hi}}])$
        \If{no collision with any $(x_j, y_j, w_j, d_j) \in \text{placed}$} \Comment{AABB check}
            \State $\mathcal{Q}[o_i] \leftarrow (x,\, y,\, p_{\max}^z + h_i/2)$
            \State Append $(x, y, w_i/2, d_i/2)$ to \text{placed};\quad $\text{found} \leftarrow \text{True}$;\quad \textbf{break}
        \EndIf
    \EndFor
    \If{$\lnot\,\text{found}$}
        \Return \textsc{Fail} \Comment{Exceeded $K_{\mathrm{try}}{=}2000$ attempts}
    \EndIf
\EndFor
\Return $\mathcal{Q}$
\end{algorithmic}
\end{algorithm}

Tasks that pass static filtering have their computed placements $\mathcal{Q}$ cached for the subsequent physics stage.

\paragraph{Stage~2: Physics Verification.}
Tasks that pass static filtering are loaded into the GRUTopia~\cite{wang2024grutopia} physics simulator (Isaac Sim backend) to validate that objects remain stable after spawning.
Each object is placed at its cached position $\mathcal{Q}[o_i]$ with a small upward offset of $+0.1$\,m, and the simulation runs for $N_{\text{settle}} = 500$ physics steps ($\Delta t = 1/240$\,s) to allow objects to come to rest.
A task is considered valid if every object satisfies:
\begin{equation}
    z_{\text{final},i} \geq z_{\text{init},i} - \epsilon_{\text{fall}},
    \qquad
    \|\mathbf{p}_{\text{final},i}\|_\infty \leq 50\,\text{m},
\end{equation}
where $\epsilon_{\text{fall}} = 0.3$\,m is the fall threshold.
Tasks passing both stages are saved with their final resting positions $\{(x_i^*, y_i^*, z_i^*)\}$, guaranteeing full reproducibility during trajectory generation.

% ------------------------------------------------------------
\subsection{Oracle Trajectory Generation}
\label{sec:supp_traj_gen}

Given a validated task episode, we execute its oracle \texttt{execution\_plan} inside the physics simulator to record a demonstration trajectory suitable for SFT.
The oracle plan is a deterministic sequence of high-level actions (pre-computed by the task generation stage); the trajectory generator executes each action and records the resulting visual observations to form action--observation pairs.

\paragraph{Environment Setup.}
At the start of each episode, all target and distractor objects are spawned at their verified positions using the world graph, the simulation undergoes a brief physics warmup to allow objects to settle, and an RGB camera mounted atop the robot's head (resolution $640 \times 480$) is initialized to provide the visual observations consumed by the policy.

\paragraph{Action Execution.}
\begin{algorithm}[t]
\caption{Oracle Trajectory Generation for One Episode}
\label{alg:traj_gen}
\small
\begin{algorithmic}[1]
\Require Episode $e$ with \texttt{execution\_plan}, \texttt{initial\_world\_graph}
\Ensure Trajectory $\tau$ (list of action--obs pairs)
\State Spawn objects per \texttt{initial\_world\_graph}; run physics warmup
\State $\tau \leftarrow [\{$\texttt{task}: $e.$\texttt{task\_description}$\}]$;\quad $\text{inv} \leftarrow \varnothing$
\For{each step $s$ in \texttt{execution\_plan}}
    \If{$s.$\texttt{action} $=$ \texttt{ask}}
        \State Append dialogue turns to $\tau$ \hfill\Comment{Social annotation (Sec.~\ref{sec:supp_social})}
    \ElsIf{$s.$\texttt{action} $=$ \texttt{nav\_to}}
        \State Find non-occluded camera position via \textsc{NavManager}
        \State Move camera; step $20$ times; append \texttt{nav\_to} + \texttt{obs(RGB)} to $\tau$
    \ElsIf{$s.$\texttt{action} $=$ \texttt{pick}}
        \State $\mathit{img}, M \leftarrow$ \textsc{ShowObjectByCategory}($s.$\texttt{target\_cat.}) \Comment{Always probe visibility first}
        \State Append \texttt{show\_object\_by\_category} $+$ \texttt{obs}($\mathit{img}$) to $\tau$
        \If{target $\in$ values($M$)} \Comment{Target visible in head-camera view}
            \State $m^* \leftarrow$ marker key of target in $M$
            \State \textsc{GazeAt}($m^*$); step $10$ times; append \texttt{gaze\_at} $+$ \texttt{obs(RGB)} to $\tau$
            \State Append \texttt{pick}($m^*$); teleport obj away; update $\mathcal{W}$; $\text{inv} \leftarrow$ target
        \Else \Comment{Target occluded: fall back to randomized walk-around}
            \State $C \leftarrow$ same-category objects at $\mathit{landmark}$ (from $\mathcal{W}$)
            \State Append \texttt{walk\_around} $+$ \texttt{obs(text enumeration of $C$)} to $\tau$
            \State \textsc{RandomShuffle}($C$) \Comment{Randomize order for trajectory diversity}
            \For{each candidate $c \in C$}
                \State \textsc{GazeAt}($c$); step $10$ times; append \texttt{gaze\_at} $+$ \texttt{obs(RGB)} to $\tau$
                \If{$c =$ target}
                    \State Append \texttt{pick}($c$); update $\mathcal{W}$; $\text{inv} \leftarrow c$; \textbf{break}
                \EndIf
            \EndFor
        \EndIf
    \ElsIf{$s.$\texttt{action} $=$ \texttt{place}}
        \State \textsc{ShowReceptacles}; overlay receptacle markers; find target marker
        \State Compute placement via \textsc{ClampAlgorithm}($\mathcal{B}_{\text{surf}}$, robot pos)
        \State Teleport $\text{inv}$ to placement; step $1000$ times; $\text{inv} \leftarrow \varnothing$
        \State Append \texttt{show\_receptacles} + \texttt{obs(RGB)} + \texttt{place} to $\tau$
    \ElsIf{$s.$\texttt{action} $\in \{$\texttt{open}, \texttt{close}$\}$}
        \State Drive door to $90^\circ$ / $0^\circ$; step $100$ times
        \State Append \texttt{open}/\texttt{close} + \texttt{obs(RGB)} to $\tau$
    \EndIf
\EndFor
\State Save $\tau$ to \texttt{.pkl}; save task metadata to \texttt{.json}
\end{algorithmic}
\end{algorithm}

\Cref{alg:traj_gen} details the oracle planner, which generates successful trajectories from the task configuration and the current simulator state.
The \texttt{pick} action deserves special attention: the oracle \emph{always} performs an object-category detection step first to probe what is currently visible from the robot's head-mounted camera.
If the target object appears in the resulting marker map, the robot gazes directly at it (\emph{visible} branch).

Otherwise it falls back to an \emph{walk\_around} search: it enumerates all same-category objects at the current landmark as text, then iterates through them in \emph{randomized} order---shuffling the exploration sequence to inject trajectory diversity rather than always visiting objects in a fixed spatial order.

% ------------------------------------------------------------
\subsection{Social Interaction Annotation}
\label{sec:supp_social}

Raw oracle trajectories contain no communicative interactions.
We augment each trajectory with realistic human--robot dialogue by injecting \texttt{ask} steps via an LLM-based annotator.

\paragraph{Annotation Strategies.}
The LLM annotator applies up to two interaction strategies per task, selected based on scene context.
\textbf{Strategy~A} (\emph{Proactive Clarification}, \texttt{ask}) is triggered when the task description is ambiguous or underspecified: a multi-turn dialogue is injected before execution in which the robot requests clarification and the user responds with additional details.
\textbf{Strategy~B} (\emph{Object Disambiguation}, \texttt{ask}) is triggered when multiple instances of the same furniture class appear in the scene (\eg two \emph{cabinets}): the robot asks which specific instance the user intends before proceeding.

\paragraph{Annotation Prompt.}
The annotator receives (i) the scene context (room layouts, furniture captions, object descriptions, and optionally a top-down scene image) and (ii) a task summary with the execution plan annotated with step indices.
The LLM annotator is prompted to reason about which strategies apply and then output a structured JSON list of annotation actions.
The key prompt template is shown below:
\promptcardfile[title={Annotation Prompt}]{prompts/social/social_annotation.txt}

\paragraph{Post-processing.}
Annotation actions are applied in reverse insertion-point order to preserve index validity.
Processing is batched asynchronously (batch size 10), with checkpointing to resume from partial runs.

\paragraph{Example Annotated Trajectory.}
Figure~\ref{fig:social_example} illustrates the effect of social annotation on a representative task.
The raw oracle plan (left) is augmented with a Strategy~B disambiguation dialogue at step~0, producing the socially-enriched trajectory (right).

\begin{figure}[h]
    \centering
    \small
    \begin{minipage}[t]{0.45\linewidth}
        \textbf{Before annotation:}
        \begin{verbatim}
0: nav_to(cabinet_1)
1: show_object_by_category(book)
2: gaze_at(book)
3: pick(book)
4: nav_to(desk_1)
5: place(desk_1)
        \end{verbatim}
    \end{minipage}
    \hfill
    \begin{minipage}[t]{0.50\linewidth}
        \textbf{After annotation (Strategy B):}
        \begin{verbatim}
0: ask [B]
   Embodied Agent: 
   "Which cabinet - the oak one
    or the glass-door one?"
   User: "The oak one, please."
1: nav_to(cabinet_1)
2: show_object_by_category(book)
3: gaze_at(book)
4: pick(book)
5: nav_to(desk_1)
6: place(desk_1)
        \end{verbatim}
    \end{minipage}
    \caption{Social annotation example. Strategy~B (disambiguation) is inserted before navigation, prompting the robot to clarify which specific furniture instance the user intends.}
    \label{fig:social_example}
\end{figure}

\subsection{Expert Trajectory Annotation}
\label{sec:supp_expert_tra_ann}

As introduced in \cref{sec:data_pipeline}, our SFT dataset requires not only the physical execution trajectories but also the internal reasoning traces and the structured historical state $h_t = (\sigma_t, q_t)$. We use \texttt{gemini-3-pro} to retroactively annotate these cognitive variables on the raw expert trajectories collected from the simulator.

\paragraph{State Alignment and Oracle Injection.}
Raw trajectories consist of paired actions $a_t$ and environmental observations $o_t^{\mathrm{env}}$ (ego-centric RGB images and text feedback). To enable \texttt{gemini-3-pro} to produce accurate visual reasoning during annotation, we inject task-specific oracle metadata into its prompt, explicitly specifying the target source object, destination receptacle, and their respective visual distractors. Crucially, this oracle information is restricted to the offline annotation phase; the final SFT dataset retains only policy-facing inputs $x_t$ free of oracle perceptual APIs to train the online policy.

\paragraph{Step-wise Reasoning Annotation.}
We initialize $a_0=\varnothing$, $\sigma_0=\varnothing$, $q_0=\operatorname{Init}(T)$, and $h_0=(\sigma_0,q_0)$. For each decision step $t=1,\ldots,L$, the annotator receives the current environmental observation $o_t^{\mathrm{env}}$, the previously executed action $a_{t-1}$, and the preceding historical state $h_{t-1}$, together with oracle metadata available only during offline annotation. Conditioned on this input, \texttt{gemini-3-pro} outputs a structured JSON containing: (1)~the reasoning trace $c_t$ supporting the next ground-truth action, (2)~a compact \texttt{History} field $\sigma_t$ summarizing tool-verified outcomes relevant to future decisions, (3)~a task-phase analysis $q_t$, and (4)~the next ground-truth action $a_t$. The two persistent fields form $h_t=(\sigma_t,q_t)$.

\paragraph{History Validation and Propagation.}
At each step, a deterministic script validates the model-produced \texttt{History} field against the recorded simulator outcome and stores the validated result as $\sigma_t$. Only the current compact state $h_t$ is propagated to step $t+1$; the raw transcript is not replayed. This construction aligns the history with the physical execution while matching the policy interface used at training and inference time.

\paragraph{Formatting for Supervised Fine-Tuning.}
The final SFT dataset uses the same single-turn-per-step format used at inference (\cref{subsec:supp_finetuned_agent}). At each step, the prompt is the policy context $x_t$ defined in \cref{eq:policy_context}, including the tool APIs $\mathcal{A}$ and the current dynamic fields. The target is the serialized response $y_t$ defined in \cref{eq:serialized_response}, which contains the reasoning trace, historical state, and action. This makes the training and inference interfaces identical.
  \FloatBarrier
  \setcounter{figure}{0}%
  \setcounter{table}{0}%
  \setcounter{algorithm}{0}%
  \section{Environment Implementation}
\label{sec:supp_tool_impl}

\subsection{Overview}
\label{sec:supp_mcp_bridge}

\begin{figure}[!t]
  \centering
  \includegraphics[width=.72\linewidth]{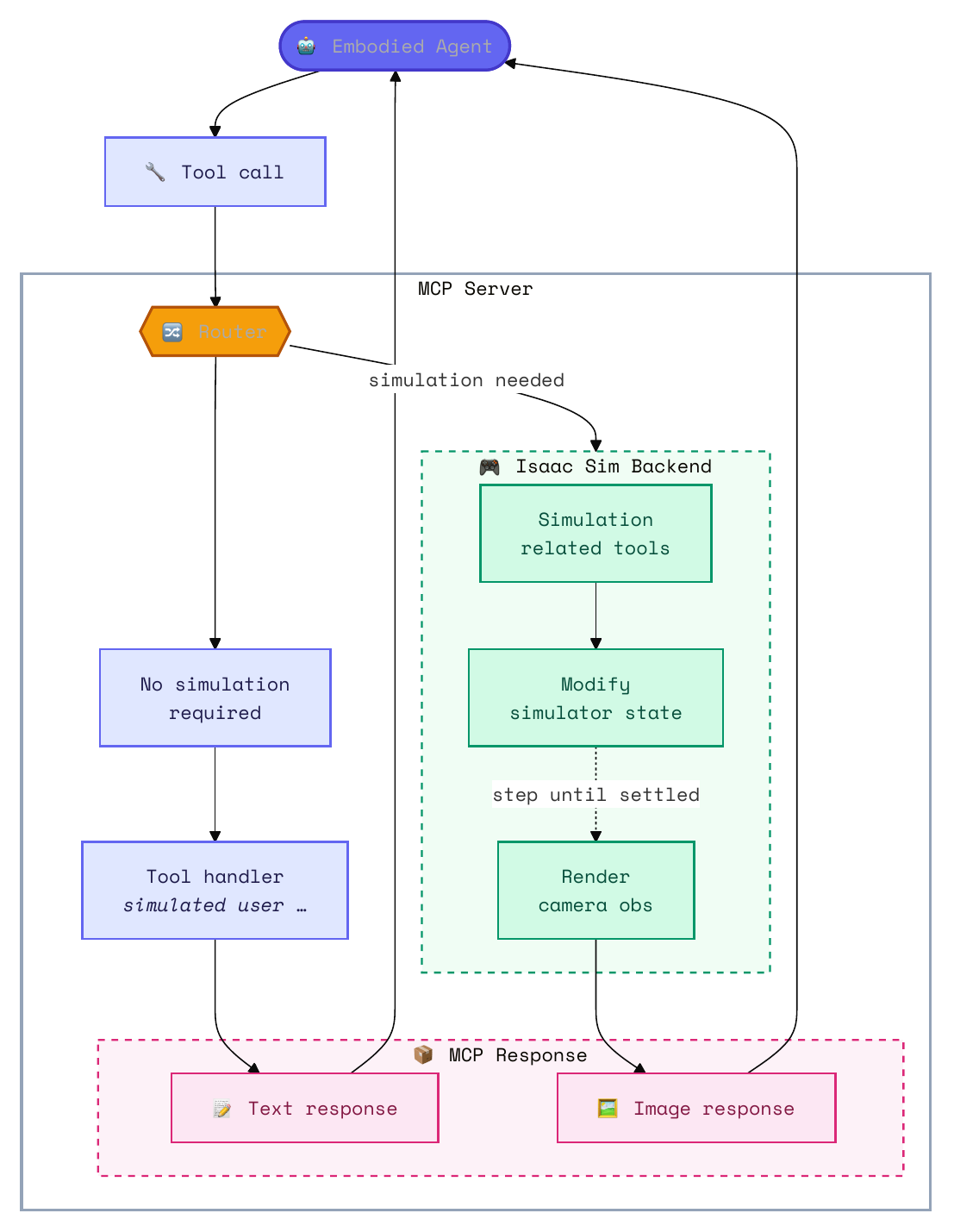}
  \caption{
    \textbf{MCP tool dispatch pipeline.}
    The embodied agent always issues the same tool calls through the same MCP schema.
    In simulation, deterministic handlers update Isaac Sim and return RGB/text observations efficiently.
    On the physical robot, the backend is swapped for onboard navigation, open-vocabulary perception, VLA manipulation, and HRI modules.
    Only the backend implementation changes; the policy-facing tool names, input signatures, and response types remain invariant.
  }
  \label{fig:supp_arch}
\end{figure}

Our system is built around the \textit{Model Context Protocol} (MCP)~\cite{anthropic2024mcp}, a standardized interface through which the embodied agent issues all tool calls and receives all responses.
As illustrated in \cref{fig:supp_arch}, each tool call first reaches a router that classifies it and dispatches it along one of two paths.
Simulation-related tools—those that require physical grounding, such as navigation, pick-and-place, and articulation—are forwarded to the Isaac Sim backend, which modifies the simulator state, steps the physics engine until the scene settles, and renders the resulting camera observation as an image response.
Tools that require no physics interaction—such as querying the user—bypass the simulation backend entirely and are handled by the simulated-user module, returning a plain-text response.
In both cases the agent receives its result through the same MCP interface, forming a closed perception–action loop without any awareness of which backend produced the response.

The same MCP interface is used when the agent operates on a physical robot. 
The router and the tool protocol remain identical; only the backend changes—the Isaac Sim engine is replaced by the robot's onboard navigation stack and manipulation system, while the MCP server continues to surface the same set of tool names, input signatures, and response types.
The agent therefore interacts with simulation and with the real robot in exactly the same way, with no modification to its tool calls or its planning logic.

\subsection{Tool Implementation Details}
\label{sec:supp_tool_impl_per}

We describe each tool in the policy-facing toolchain, which is free of oracle perceptual APIs (see \cref{sec:environment}), along with its concrete simulation implementation.

\paragraph{Sim-to-Real Rationale: Simulation Efficiency versus Physical Fidelity.}
The core design of our system resolves the tension between simulation efficiency and physical fidelity by separating the policy-facing interface from the backend implementation. During simulation, deterministic server-side handlers execute navigation, perception, and manipulation effects efficiently, reducing trajectory noise and per-step wall-clock time for large-scale data collection. Crucially, these handlers do not expose oracle perceptual APIs to the policy: the agent receives only RGB/SoM images, text responses, execution tracebacks, and the same tool return format used at deployment. This preserves the invariant required for zero-shot transfer. In the real world, only the backend changes: deterministic simulation handlers are replaced by the robot's navigation stack, open-vocabulary perception, VLA manipulation model, and HRI interface, while the agent weights, tool names, input signatures, and response types remain unchanged.

\paragraph{Physical Infrastructure.}
To enable this deployment on the ARX LIFT2 platform and bridge the sim-to-real gap, we integrate ConceptGraph~\cite{gu2024conceptgraphs} with an Intel RealSense D455 camera for semantic 3D scene graph construction. Furthermore, to overcome the RealSense camera's relocalization errors exceeding 5\,cm, we introduce a dual-tracking architecture. This setup features an Odin1 depth camera with proprietary SLAM, effectively reducing relocalization error to under 1\,cm. It is coupled with a robust navigation stack using A* search for global planning, the Dynamic Window Approach (DWA) for local obstacle avoidance, and PID controllers to ensure reliable, fine-grained mobility.

\paragraph{\texttt{nav\_to}~(inter-receptacle navigation).}
\textbf{Simulation.}
Given a target receptacle name, the handler identifies a collision-free camera position that maintains an unobstructed line of sight to the primary interaction surface of the receptacle (\eg the top shelf or door).
To do so, it determines the axis-aligned bounding box of the navigable part of the receptacle, derives an appropriate standoff radius, and queries an occupancy-map-based navigation module for the closest non-occluded vantage point on the two-dimensional floor plan.
The observation is rendered from the new camera pose, and the tool returns the RGB render of the resulting viewpoint.
\textbf{Real World.}
The MCP server queries the target node from the ConceptGraph, and an A* planner generates a trajectory from the current pose of the robot to a valid interaction point.
The goal pose maintains a standoff distance of at least 0.6 meters from the target to ensure sufficient manipulation workspace.
If no valid trajectory exists, the tool returns a failure log.
The robot tracks this path using DWA and PID control, returning the final RGB frame upon arrival.

\paragraph{\texttt{walk\_around}~(intra-receptacle broad scanning).}
\textbf{Simulation.}
The simulation maintains a per-receptacle object registry that is kept synchronized throughout each episode: every \texttt{pick} operation removes an object entry from the source receptacle, and every \texttt{place} operation inserts an entry at the destination, ensuring the registry always reflects the current scene state.
When \texttt{walk\_around} is invoked, the handler reads this live registry for the active receptacle of the agent and resolves each entry to the corresponding semantic category, returning a natural-language inventory string (\eg \textit{``I found the following object(s): 0: a(an) apple.''}) paired with an integer-to-category index for downstream reference.
\textbf{Real World.}
The implementation generates predefined navigation waypoints covering the entire field of view of the occupancy map for each furniture node.
The robot traverses these waypoints and captures images, which Grounded-SAM-2~\cite{ren2024grounded} processes to extract object masks.
These masks are deduplicated across viewpoints using visual feature matching and are subsequently merged into an annotated image.
Following this, the robot navigates to the waypoint closest to the cluster center of the target object for subsequent interactions.

\paragraph{\texttt{show\_object\_by\_category}.}
\textbf{Simulation.}
This tool implements open-vocabulary object detection through a four-stage perception pipeline.
First, the photorealistic renderer of the simulator is coupled with a per-instance segmentation annotator to produce pixel-accurate object masks for all entities visible in the current view.
Second, text embeddings are computed for both the open-vocabulary query of the agent and every asset category present in the scene using an embedding service; the category with the highest cosine similarity to the query is selected as the match.
Third, pixel-level masks for the matched instances are extracted from the segmentation output.
Fourth, numbered Set-of-Mark~\cite{yang2023setofmark} overlays—colored region masks labeled with integer identifiers—are composited onto the RGB image.
The resulting annotated image is returned to the agent, and the id-object mapping is stored in the environment for parsing following tool calls that reference the integer identifiers. 
\textbf{Real World.}
The perception interface is replicated in the physical deployment using Grounded-SAM-2~\cite{ren2024grounded}.
The model receives the queried category name, performs open-vocabulary segmentation, and assigns an integer ID to each generated mask.
These specific IDs and masks subsequently serve as the visual prompts fed into the downstream VLA model.

\paragraph{\texttt{show\_receptacles}~(receptacle labeling).}
\textbf{Simulation.}
The handler retrieves the 3D bounding boxes of all receptacles within the current camera frustum, projects each box into the image plane to compute a 2D region mask, and composites numbered Set-of-Mark~\cite{yang2023setofmark} overlays onto the RGB frame.
The annotated image together with the integer-to-receptacle-name mapping is returned to the agent, enabling subsequent \texttt{nav\_to} or \texttt{place} calls to reference receptacles by the corresponding visual identifiers.
\textbf{Real World.}
In the physical deployment, the annotations are registered against the head-mounted RGB-D point cloud and the ConceptGraph to align 3D bounding boxes with the 2D masks.
Similar to the object detection pipeline, the specific IDs and masks function as visual prompts for the downstream VLA model.

\paragraph{\texttt{gaze\_at}~(target approach and alignment).}
\textbf{Simulation.}
This tool moves the robot close to a specified object for detailed observation and subsequent grasping.
The target object may originate from two upstream tools with different output modalities: \texttt{walk\_around} returns a 3D bounding box for each discovered object, whereas \texttt{show\_object\_by\_category} yields a 2D segmentation mask from which a 3D centroid is recovered via depth-camera back-projection.
Given the 3D position of the object, the handler queries the occupancy map for a collision-free approach pose within the reach of the robotic arm, commands the mobile base to navigate to that pose, and reorients the camera toward the object.
In simulation, the 3D positions are read directly from the scene graph.
When marker persistence is requested, the Set-of-Mark overlays are re-rendered in the new close-up view, ensuring that visual identifiers remain consistent across viewpoint changes for subsequent grasping calls made by the agent.
\textbf{Real World.}
If the target mask identified by the associated ID is not clearly visible in the current field of view, the system retrieves the navigation waypoint from the \texttt{walk\_around} history that originally captured the mask.
The robot autonomously navigates back to this optimal viewpoint to center the target for manipulation.

\paragraph{\texttt{pick}~/~\texttt{place}~/~\texttt{open}~/~\texttt{close}~(manipulation primitives).}
\textbf{Simulation.}
All four manipulation tools follow the same pattern: the handler resolves the integer marker provided by the agent to the corresponding object or receptacle in the scene graph and subsequently executes a deterministic state change.
Specifically, \texttt{pick} removes the target object from the scene and transfers the item to a virtual inventory; \texttt{place} restores the held object onto a designated surface at a collision-free position computed from the bounding box of the receptacle; \texttt{open} and \texttt{close} drive the articulated joint of a door or drawer to the fully open or resting angle, respectively.
Each operation updates the per-receptacle object registry and the state of the world graph accordingly, while the simulator re-renders an RGB observation from the resulting scene.
\textbf{Real World.}
A fine-tuned Vision-Language-Action (VLA) model executes all manipulation primitives.
Because Grounded-SAM-2~\cite{ren2024grounded} functions as the dataset preprocessing scheme during the visually-instructed fine-tuning phase of the VLA, the deployed VLA model directly ingests the mask and ID generated by the perception tools as the initial visual prompt.
This architecture effectively maps the RGB-D observation to a 6-DoF end-effector trajectory.

\paragraph{\texttt{ask}~(natural language query to user).}
\textbf{Simulation.}
This tool bypasses the physics dispatch pipeline and is routed directly to the Simulated User module.
The Simulated User is a stateful language-model wrapper that maintains a task-specific system prompt encoding the goal state and the preference profile of the user, an interleaved conversation history, and a synchronized view of the most recent action and observation of the agent.
When the agent invokes \texttt{ask}, the module injects the current state of the robot into the system prompt, appends the question to the conversation history, and queries a language-model backend to produce a reply.
The response is extracted from a structured output format and returned to the agent as a plain-text tool result, rendering the text indistinguishable in form from the output of any other tool.
\textbf{Real World.}
For physical deployment, we developed a Human-Robot Interaction GUI.
To ensure the human user can act as an oracle substitute, the GUI displays the decision history of the agent, reasoning processes, and live egocentric observations.
The human user types a response based on this context, effectively closing the loop for physical execution.

\subsection{Simulated User Module}
\label{subsec:supp_simuser}

The simulated user is driven by \texttt{gemini-3-flash} and serves as an interactive collaborator that answers questions from the embodied agent.
To replicate realistic human behavior, the simulated user is governed by two complementary mechanisms: a \emph{profile} that defines its persona and behavioral constraints, and an \emph{observation interface} that keeps it synchronized with the evolving state of the robot.

\paragraph{Profile template.}
Each task instance initializes the simulated user with a structured profile containing three elements: the initial object layout $\mathcal{W}_\text{init}$, the goal layout $\mathcal{W}_\text{goal}$, and a fuzzy natural-language instruction that deliberately omits the exact target identity.
To ensure that the simulated user behaves like a real human collaborator—goal-directed yet informationally constrained—behavioral rules are embedded in the profile:
\begin{enumerate}
    \item \textbf{Goal-driven.} Every response must guide the robot from $\mathcal{W}_\text{init}$ toward $\mathcal{W}_\text{goal}$.
    \item \textbf{Information barrier.} The user must \emph{not} reveal the exact position of the target object; only vague references such as ``check what's inside'' are permitted.
    \item \textbf{Dynamic clarification.} The user may name the specific target object \emph{only after} the robot explicitly reports seeing it in its current observation, faithfully simulating how a real person progressively discloses information as the situation unfolds.
    \item \textbf{No raw IDs.} Internal alphanumeric object identifiers present in $\mathcal{W}_\text{init}$ and $\mathcal{W}_\text{goal}$ are never spoken verbatim; the user maps them to natural-language descriptions derived from the fuzzy instruction or visual captions.
\end{enumerate}

\noindent Together, rules 2 and 3 create a progressive disclosure loop: the user initially withholds specifics, the robot explores and reports what it perceives, and the user then grounds its follow-up guidance in those concrete observations.
Below is an illustrative profile for a ``move eating utensil'' task:

\promptcardfile[title={Example User Profile}]{prompts/social/social_agent_profile.txt}

\paragraph{Auxiliary observation.}
A fundamental representation gap exists between the simulated user and the embodied agent: the user profile encodes object layouts using internal alphanumeric IDs (\eg \texttt{\_4004e25c\ldots}), whereas the agent perceives the world exclusively through ego-centric RGB images.
These two modalities are incommensurable—the simulated user cannot directly ``see'' what the robot's camera captures, nor can it infer from raw IDs alone which objects have entered the agent's field of view.
To bridge this gap and allow the simulated user to accurately track the agent's exploration progress, we introduce an auxiliary textual observation.

After each tool call, the environment performs a visibility check on the current render and composes a structured summary that records (i)~the action the robot just executed, and (ii)~a list of objects now visible in its ego-centric view, each described by its semantic category and a natural-language caption.
This summary replaces the previous one in the simulated user's context at every step, ensuring that it always reasons over the most recent robot state without accumulating a growing history.
An example is shown below:

\promptcardfile[title={Auxiliary Observation}]{prompts/social/obs_injection.txt}

\noindent By expressing visible objects through semantic captions (\eg ``a pair of wooden chopsticks with red tips'') rather than raw asset identifiers, the auxiliary observation serves two purposes.
First, it enables the simulated user to apply the \emph{no-raw-IDs} rule naturally, since the captions provide exactly the kind of everyday language a real person would use.
Second, it closes the progressive disclosure loop defined by rules~2 and~3: upon recognizing that a semantically described object matches the fuzzy instruction (\eg that ``wooden chopsticks'' qualifies as the requested ``eating utensil''), the simulated user can unambiguously name the target in its next response.
  \FloatBarrier
  \setcounter{figure}{0}%
  \setcounter{table}{0}%
  \setcounter{algorithm}{0}%
  \section{Reinforcement Learning Setup}
\label{app:env_pomdp}

\subsection{Full POMDP Definition}
\label{app:pomdp_definition}
We provide the complete formalization of the single-step decision-making process. The embodied agent operates within a Partially Observable Markov Decision Process (POMDP) defined by the tuple $\mathcal{M}=\langle \mathcal{S}, \mathcal{A}, \mathcal{O}, \mathcal{P}, \mathcal{R}, \Omega, \gamma \rangle$. 

The unobservable state space $\mathcal{S}$ contains precise 3D object poses, the physical World Graph, and the internal simulator dynamics. The action space $\mathcal{A}$ contains 11 non-privileged tools spanning navigation, perception, manipulation, communication, and episode termination. The environmental observation space is $\mathcal{O}$, with $o_t^{\mathrm{env}}\sim\Omega(\cdot\mid s_t)$ as in \cref{eq:env_observation}; $o_t^{\mathrm{env}}$ contains ego-centric RGB views, execution tracebacks, and dynamic tool returns. The policy augments this observation with the preceding action and its compact memory to form $x_t$ in \cref{eq:policy_context}. These agent-maintained variables are part of the policy context rather than outputs of the environment observation kernel.

After the policy selects $a_t$, the environment evolves according to
\begin{equation}
s_{t+1}\sim\mathcal{P}(\,\cdot\mid s_t,a_t),
\qquad
r_t=\mathcal{R}(s_t,a_t,s_{t+1}).
\label{eq:pomdp_transition_reward}
\end{equation}
The transition includes both physical changes executed by Isaac Sim and conversational responses from the simulated user. The discount factor is $\gamma=0.99$.

\subsection{Action Space and Tool Configurations}
\label{app:action_space}
The policy model outputs a reasoning block followed by a JSON object that specifies the updated history $h_t$ and the selected tool call $a_t \in \mathcal{A}$. The environment enforces physical constraints during execution. Episodes terminate if the agent outputs the \texttt{finish} action, reaches the maximum horizon of $H_{\max}=30$ decision steps, or encounters an unrecoverable error.

\subsection{Detailed Reward Configuration}
\label{app:reward}
Building upon \cref{sec:reward_design}, the transition reward is the sum of signed components:
\begin{equation}
r_t =
r_t^{\mathrm{task}}+
r_t^{\mathrm{prog}}+
r_t^{\mathrm{phys}}+
r_t^{\mathrm{ask}}+
r_t^{\mathrm{time}}+
r_t^{\mathrm{safe}}.
\label{eq:reward_decomposition}
\end{equation}

We first define the goal-change set and then augment it with execution-specific targets:
\begin{equation}
\Delta_{\mathrm{goal}}
=\text{WG}_{\mathrm{goal}}\ominus\text{WG}_{\mathrm{init}},
\qquad
\Delta_{\mathrm{task}}
=\Delta_{\mathrm{goal}}\cup\Delta_{\mathrm{aux}}.
\label{eq:reward_target_sets}
\end{equation}
Here $\Delta_{\mathrm{aux}}$ contains navigation and articulation targets used only for shaping. Let $\eta_t$ denote the auxiliary events triggered by $a_t$ (\eg a first visit to the source receptacle). The observed event set and normalized shaping progress are
\begin{equation}
\delta_t=
\left(\text{WG}_{t+1}\ominus\text{WG}_t\right)\cup\eta_t,
\qquad
\rho_t=1-\frac{|\Delta_{\mathrm{rem},t}|}{|\Delta_{\mathrm{task}}|}.
\label{eq:task_progress}
\end{equation}
Here $\Delta_{\mathrm{rem},t}\subseteq\Delta_{\mathrm{task}}$ is the target set remaining before action $a_t$. For each newly matched element of $\delta_t\cap\Delta_{\mathrm{rem},t}$, that element is removed and $r_t^{\mathrm{task}}$ receives $+1.5$; completed elements cannot receive duplicate credit. We use the progress-difference term $r_t^{\mathrm{prog}}=2.0(\rho_{t+1}-\rho_t)$. When all goal-state targets are complete, $r_t^{\mathrm{task}}$ receives a one-time $+2.0$ bonus. Within $r_t^{\mathrm{phys}}$, reversing a previously achieved physical state incurs $-0.5$, and placing an object on an incorrect receptacle incurs $-2.0$.

While articulation actions and navigation primitives do not directly modify the core object--receptacle relations, they are essential for execution. Their targets in $\Delta_{\mathrm{aux}}$ (\eg a first visit to \texttt{nav\_to(src)} or the attribute change \texttt{(receptacle, open)}) use the same matching logic and $+1.5$ reward as object-manipulation targets. The physics engine enforces temporal constraints such as opening a receptacle before placing an object inside it.

To alleviate sparse rewards, $r_t^{\mathrm{phys}}$ assigns $+0.3$ to a valid pick, a one-time $+0.5$ credit when the policy enters a valid in-transit holding state, and $+1.0$ for recovery after an incorrect pick. A valid disambiguation query receives $r_t^{\mathrm{ask}}=+1.0$, subject to the difficulty-conditioned budget $B_{\mathrm{ask}}\leq3$.

We set the positive coefficient $\lambda_{\mathrm{time}}=0.03$ and apply $r_t^{\mathrm{time}}=-\lambda_{\mathrm{time}}$ after the initial three decisions. Safety is evaluated from the traceback returned in $o_{t+1}^{\mathrm{env}}$: severe errors, such as an invisible target bounding box or invalid API format, set $r_t^{\mathrm{safe}}=-0.5$, while moderate perception errors and gripper-constraint violations set it to $-0.3$.

\subsection{GSPO Objective and Distributed Training}
\label{app:training}
For each prompt, the rollout policy $\pi_{\theta_{\mathrm{old}}}$ samples a group of $G$ interactive trajectories. Let $z_i=(z_{i,1},\ldots,z_{i,|z_i|})$ denote the concatenated model-generated tokens in rollout $i$, with token context $\xi_{i,k}$, and let $J_i=\sum_{t=1}^{L_i}\gamma^{t-1}r_{i,t}$ be its discounted return. We define the group-relative advantage and the length-normalized sequence ratio as
\begin{equation}
\widehat{A}_i=
\frac{J_i-\overline{J}}
{\operatorname{Std}(J_1,\ldots,J_G)+\epsilon_A},
\qquad
s_i(\theta)=
\exp\!\left[
\frac{1}{|z_i|}
\sum_{k=1}^{|z_i|}
\log
\frac{\pi_\theta(z_{i,k}\mid\xi_{i,k})}
{\pi_{\theta_{\mathrm{old}}}(z_{i,k}\mid\xi_{i,k})}
\right],
\label{eq:gspo_ratio}
\end{equation}
where $\overline{J}=G^{-1}\sum_{i=1}^{G}J_i$ and $\epsilon_A>0$ prevents division by zero. Let $\kappa_i(\theta)$ denote the length-normalized token-level KL divergence from the fixed SFT reference policy:
\begin{equation}
\kappa_i(\theta)=
\frac{1}{|z_i|}\sum_{k=1}^{|z_i|}
D_{\mathrm{KL}}\!\left(
\pi_\theta(\,\cdot\mid\xi_{i,k})
\middle\|
\pi_{\mathrm{ref}}(\,\cdot\mid\xi_{i,k})
\right).
\label{eq:gspo_kl}
\end{equation}
GSPO minimizes
\begin{equation}
\mathcal{L}_{\mathrm{GSPO}}=
-\frac{1}{G}\sum_{i=1}^{G}
\min\!\left(
s_i(\theta)\widehat{A}_i,\,
\operatorname{clip}\!\left(s_i(\theta),1-\epsilon,1+\epsilon\right)
\widehat{A}_i
\right)
+\frac{\beta}{G}\sum_{i=1}^{G}\kappa_i(\theta),
\label{eq:gspo_objective}
\end{equation}
where $\pi_{\mathrm{ref}}$ is the fixed SFT reference policy, $\epsilon$ is the clipping radius, and $\beta$ is the KL coefficient.

The one-epoch SFT checkpoint used to initialize GSPO is derived from Qwen3-VL-8B-Instruct. For this preceding SFT stage, we use a learning rate of $5\times10^{-6}$, per-device batch size 2, gradient accumulation steps 4, warmup ratio 0.08, image resolutions from 784 to 100,352 pixels, an 8,192-token context, and bfloat16 precision on 8 NVIDIA A800 GPUs.

The training backend uses nine task pools to provide 144 concurrent Isaac Sim endpoints via MCP. A central server pool assigns an exclusive simulation instance to each prompt group (group size $G=8$) to maintain consistency for the computation of the GSPO advantage $\widehat{A}_i$. The optimization of the model runs on a single 8-GPU NVIDIA 4090 node to ensure compatibility with Isaac Sim. Two GPUs host the vLLM rollout server for trajectory generation, and the remaining six GPUs run the trainer using DeepSpeed ZeRO-2 with 6-way data parallelism. The hyperparameters for training are listed in \cref{tab:hyperparameters}.

\begin{table}[h]
\centering
\caption{GSPO Training Hyperparameters}
\label{tab:hyperparameters}
\begin{tabular}{ll}
\toprule
Hyperparameter & Configuration \\
\midrule
Base Model & Qwen3-VL-8B-Instruct (SFT Checkpoint) \\
Parameter Efficient Tuning & LoRA (rank 32, $\alpha = 64$, target: all linear layers) \\
Vision Encoder & Frozen \\
Optimizer & AdamW (via DeepSpeed ZeRO-2) \\
Learning Rate & $5 \times 10^{-6}$ \\
KL Penalty Coefficient ($\beta$) & 0.04 \\
GSPO Group Size ($G$) & 8 \\
Per-Device Batch Size & 1 \\
Gradient Accumulation Steps & 8 \\
Effective Unique Prompts per Step & 6 \\
Max Training Steps & 1000 \\
\bottomrule
\end{tabular}
\end{table}

\subsection{Evaluation Metrics}
\label{app:metrics}
The performance of the system is measured across five primary dimensions. Let $\Delta_{\mathrm{goal,rem,term}}\subseteq\Delta_{\mathrm{goal}}$ denote the goal-state changes still unsatisfied at termination. The success rate (SR) is the proportion of episodes with $|\Delta_{\mathrm{goal,rem,term}}|=0$. The completion rate (CR), defined as $1-|\Delta_{\mathrm{goal,rem,term}}|/|\Delta_{\mathrm{goal}}|$, quantifies partial goal completion without counting auxiliary shaping targets. Cumulative reward is the undiscounted evaluation statistic $\sum_t r_t$. Episode length measures the number of decision steps, and ask frequency is the average number of communicative tool invocations per episode.

\subsection{Analysis of RL Training: Successes and Bottlenecks}
\label{sec:rl_analysis}

We analyze 7,304 rollout trajectories generated during the RL phase to evaluate the impact of GSPO training. The overall in-training success rate converged to 39.3\%. Beyond numerical metrics, the analysis of the trajectories reveals two distinct trends: the agent demonstrates improved logical reasoning, yet the performance of the agent is constrained by hardware-induced perceptual bottlenecks.

\paragraph{Cognitive and Interactive Policy Improvements}
The RL phase induced cognitive capabilities that were unstable in the SFT baseline, allowing the agent to construct reliable pipelines and resolve open-world uncertainty. The agent developed a canonical execution flow: \texttt{nav\_to} $\rightarrow$ \texttt{show\_object\_by\_category} $\rightarrow$ \texttt{gaze\_at} $\rightarrow$ \texttt{pick} $\rightarrow$ \texttt{nav\_to} $\rightarrow$ \texttt{show\_receptacles} $\rightarrow$ \texttt{place}. This indicates that the agent systematically narrows down visual searches before committing to a target and surveys placement areas rather than guessing the states of the receptacles.

Furthermore, the \texttt{ask} tool, rarely invoked in the SFT dataset, appeared in 15.8\% of the RL trajectories. In 90.3\% of these instances, the agent correctly identified multiple visual distractors (e.g., ``I see a watch and a clock on the nightstand. Which one do you want?'') and paused execution to query the simulated user. This suggests that the GSPO optimization drove the policy to learn human-robot collaboration from sparse rewards.

\paragraph{Perceptual Bottlenecks.}
Despite improvements in logical planning, the in-training success rate was limited by visual recognition failures. An analysis of the 4,431 failed trajectories indicates that 94.3\% were not pipeline failures; the agent executed the sequence of \texttt{pick} and \texttt{place} correctly but manipulated the wrong object.

These hallucinated marker selections stem from fine-grained visual grounding limits under the frozen 411M-parameter ViT backbone. Both SFT and RL use \texttt{max\_pixels}$=100{,}352$ for image inputs, and a resolution-alignment control changes the final success rate by only about $\pm3\%$ across task families. Thus, the performance gains of GSPO do not arise from a resolution mismatch. Instead, the remaining errors reflect cases where visually similar distractors differ by subtle attributes such as texture, color, or part-level geometry that the frozen visual representation cannot reliably bind to the instruction.

\paragraph{Discrepancy Between Training and Benchmark Performance.}
The 39.3\% in-training success rate differs from the final evaluation on the offline benchmark primarily because training rollouts are sampled online from a heterogeneous task pool without an explicit curriculum. The intrinsic difficulty within a single GSPO sampling group fluctuates (e.g., distinct books versus highly ambiguous geometric distractors), introducing variance to the step-wise success rate. The benchmark suite, however, evaluates the agent across a fixed and more uniformly stratified task distribution, yielding a more stable assessment of the policy. Overall, GSPO optimizes the cognitive policy for reasoning and interaction, while the observed training ceiling is mainly a fine-grained perception bottleneck rather than an algorithmic flaw in the POMDP or RL formulation.
  \FloatBarrier
  \setcounter{figure}{0}%
  \setcounter{table}{0}%
  \setcounter{algorithm}{0}%
  \section{Agent Inference and Evaluation Setup}
\label{sec:supp_eval_details}

\subsection{Zero-Shot VLM Baseline}
\label{subsec:supp_zeroshot_agent}

We evaluate a family of proprietary VLMs as zero-shot embodied agents on our benchmark.
Each model is queried through an OpenAI-compatible chat completion API with native function-calling support.

At the beginning of each episode, the agent receives a system prompt that encodes the task rules, the marker-based object-reference protocol, and the available action vocabulary.
The system prompt is augmented with a \emph{scene description} that enumerates all rooms and their furniture/receptacles in the current house layout, giving the agent a static, text-only overview of the environment. The full system prompt template is shown below:

\promptcardfile[title={Zero-Shot Agent Prompt}]{prompts/zero_shot/system_prompt.txt}

Below is an example scene description that populates the \emph{scene description} placeholder:
\promptcardfile[title={Example Scene Description}]{prompts/qwen_eval/example_scene_description.txt}

The task instruction is provided as the first user message, after which the agent is expected to respond with a tool call at each step.
The tool calls are forwarded to the simulation server 
and the response (textual or visual observation) is appended to the context for the next turn.
Each episode runs for a maximum of $H_{\max} = 30$ steps.
The episode terminates when the agent calls the \texttt{finish} action or exhausts the step budget.
Success is determined by comparing the final world state against the ground-truth goal state in the task configuration.

\subsection{\realagent (Ours)}
\label{subsec:supp_finetuned_agent}

As described in Sec.~\ref{sec:data_pipeline}, our agent reformats the entire episode state into a single user turn at every step.
The four dynamic placeholders in the prompt template map directly to the policy context: \texttt{\{TASK\}} $\leftrightarrow T$, \texttt{\{TASK\_PROGRESS\}} $\leftrightarrow h_{t-1}$, \texttt{\{LAST\_ACTION\}} $\leftrightarrow a_{t-1}$, and \texttt{\{LAST\_OBS\}} $\leftrightarrow o_t^{\mathrm{env}}$.
The tool list is embedded verbatim in the prompt as a JSON array rather than fetched dynamically from the MCP server.

\promptcardfile[title={System Prompt — Fine-Tuned Embodied Agent}]{prompts/qwen_eval/qwen_embodied_agent.txt}

\promptcardfile[title={Tool Definitions (JSON Schema)}]{prompts/qwen_eval/tool_list.txt}

\paragraph{Example scene and output.}
Below is a representative model output illustrating the three-part structure (chain-of-thought, summary dictionary, and action JSON).

\promptcardfile[title={Example Model Output}]{prompts/qwen_eval/example_model_output.txt}

\paragraph{Episode loop.}
Algorithm~\ref{alg:finetuned_infer} summarises the per-episode inference procedure.
Each episode runs for a maximum of $H_{\max} = 30$ steps with greedy decoding and a budget of 4096 new tokens per step.
Only the images from the most recent action are included in the current prompt; prior images are discarded.

\begin{algorithm}[H]
\caption{Fine-tuned agent episode loop}
\label{alg:finetuned_infer}
\begin{algorithmic}[1]
\Require Task instruction $T$, scene description $\mathcal{C}$, max steps $H_{\max}$
\State Build prompt template $\mathsf{Prompt}(\mathcal{C})$ \Comment{embed scene once}
\State $a_0 \leftarrow \varnothing$,\quad $\sigma_0 \leftarrow \varnothing$,\quad $q_0 \leftarrow \operatorname{Init}(T)$
\State $h_0 \leftarrow (\sigma_0,q_0)$,\quad $o_1^{\mathrm{env}} \leftarrow \textsc{Env.Reset}()$
\For{$t = 1, 2, \ldots, H_{\max}$}
  \State $x_t \leftarrow (T,\,\operatorname{Serialize}(\mathcal{A}),\,o_t^{\mathrm{env}},\,a_{t-1},\,h_{t-1})$
  \State $p_t \leftarrow \mathsf{Prompt}(\mathcal{C})[x_t]$ \Comment{fill dynamic fields}
  \State $\hat{y}_t \leftarrow \textsc{Generate}(\text{model},\, p_t,\, \text{images})$
  \State $(c_t,\,\sigma_t,\,q_t,\,a_t) \leftarrow \textsc{Parse}(\hat{y}_t)$
  \State $h_t \leftarrow (\sigma_t,q_t)$
  \If{$a_t.\texttt{tool\_name} = \texttt{finish}$}
    \State $\textit{success} \leftarrow \textsc{Eval}(\text{current scene state}, \text{goal state})$
    \State \textbf{break}
  \EndIf
  \State $(o_{t+1}^{\mathrm{env}},\, \text{images}) \leftarrow \textsc{Env}(a_t)$
\EndFor
\end{algorithmic}
\end{algorithm}

\subsection{Simulated-User Fidelity and Human-User Generalization}
\label{subsec:supp_user_fidelity}

\paragraph{Simulated-user answer fidelity.}
To verify that the simulated user provides useful but bounded clarifications, we manually evaluate 50 answers generated by the Gemini-driven user module. Among these answers, 47 are accurate and sufficient for progressing toward the target goal, 2 are directionally correct but lack enough detail to fully disambiguate the target, and 1 is incorrect. This evaluation supports the use of the simulated user as a scalable proxy for training and benchmark interaction, while preserving the limitation that simulated answers are not perfectly reliable.

\paragraph{Generalization to real users.}
We additionally evaluate the trained agent on a subset of 50 SUL tasks under two user sources: the Gemini simulated user and a real human user. Real-user responses are substantially shorter than simulated responses (4.3 words on average versus 30.2 words), but the agent obtains comparable success counts: 27 successful episodes with the human user versus 29 with Gemini. The small drop occurs primarily when the human response does not fully distinguish the intended target among visual distractors, indicating that the policy can transfer from simulated dialogue to real users but remains sensitive to underspecified feedback.
  \FloatBarrier
  \setcounter{figure}{0}%
  \setcounter{table}{0}%
  \setcounter{algorithm}{0}%
  \section{Qualitative Analysis}
\label{sec:supp_case_study}

To complement the quantitative failure taxonomy in the main paper (\cref{sec:experiments}), we present step-by-step trajectory visualizations for representative failure and success episodes from our benchmark evaluation.
Each case traces the agent's tool calls and ego-centric observations, illustrating both the capabilities and remaining limitations of our agent.

\subsection{Failure Case Analysis}

We select four failure episodes that cover the dominant failure modes identified in the main-paper analysis (\cref{fig:failure_sankey}): \emph{Object Confusion} (including wrong object selection and similar furniture confusion), \emph{Key Action Missing}, and \emph{Lost Memory}.

\begin{figure}[!ht]
  \centering
  \includegraphics[width=\linewidth]{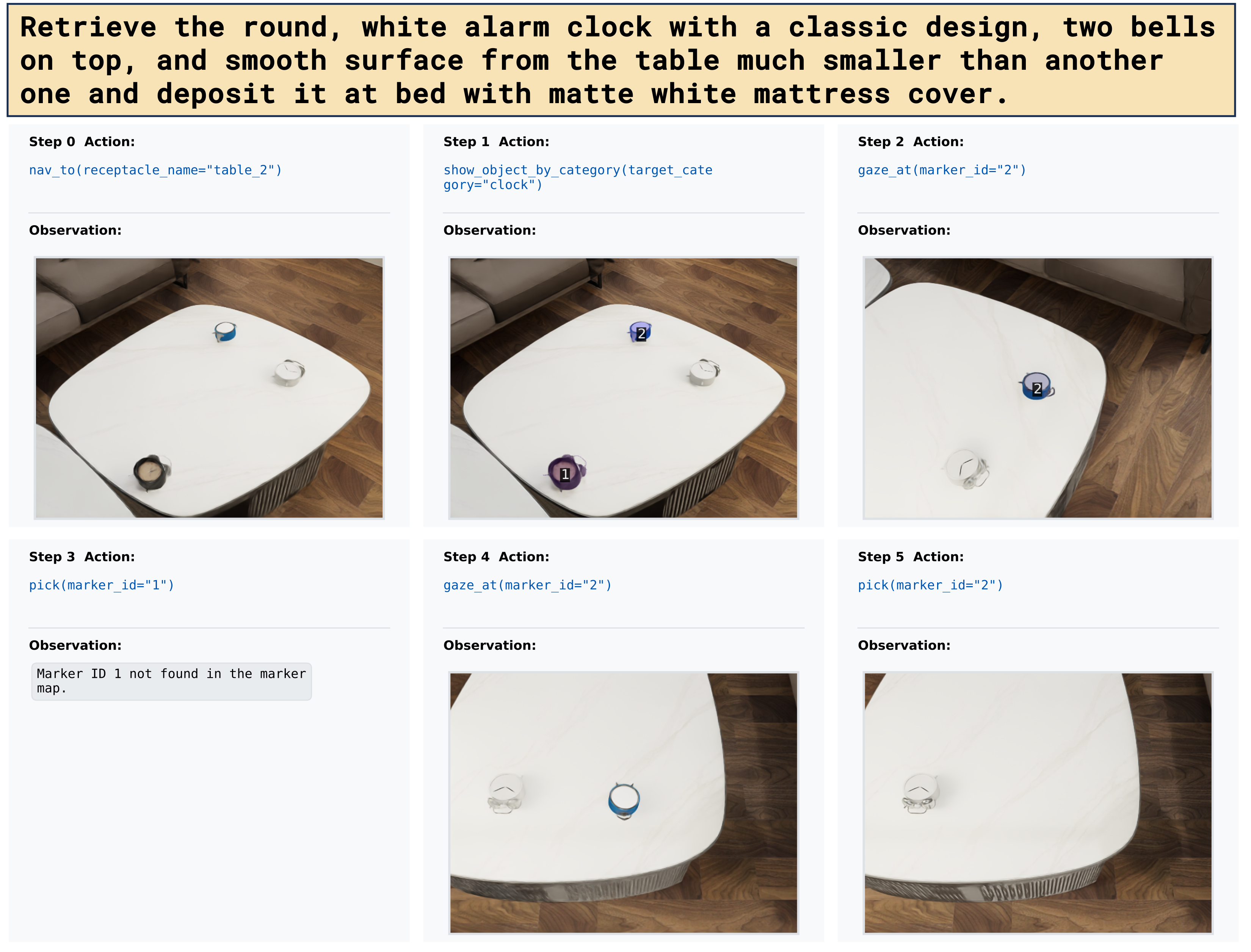}
  \caption{\textbf{Failure: Wrong object selection.} Task: retrieve a \textit{white} alarm clock from \texttt{table\_2}. After detecting two candidate clocks---one blue and one white (Step~1), the agent fails to ground the color attribute and picks the blue alarm clock instead of the target white one.}
  \label{fig:case_wrong_obj}
\end{figure}

\paragraph{Wrong Object Selection.}
\Cref{fig:case_wrong_obj} depicts an episode with the instruction: \textit{``Retrieve the round, white alarm clock with a classic design, two bells on top, and smooth surface from the table much smaller than another one and deposit it at bed with matte white mattress cover.''}
The agent navigates to \texttt{table\_2} (Step~0) and invokes \texttt{show\_object\_by\_category(target\_category="clock")} (Step~1), which produces two numbered marker overlays on candidate objects---a blue alarm clock and the target white alarm clock.
Despite the instruction explicitly specifying a \textit{white} alarm clock, the agent fails to correctly ground the color attribute and ultimately picks the blue alarm clock instead (Steps~2--5). A stale marker~ID error also occurs during this process (Step~3), but the fundamental failure is one of visual attribute discrimination: the agent recognizes both objects as clocks yet cannot reliably distinguish them based on the color and appearance attributes specified in the instruction.
This case exemplifies a limitation in fine-grained visual grounding. The RL reward penalizes the incorrect pick outcome but does not provide explicit supervision on attribute-level object matching, leaving the agent unable to resolve visually similar candidates when the distinguishing features are subtle surface properties such as color.

\begin{figure}[!ht]
  \centering
  \includegraphics[width=\linewidth]{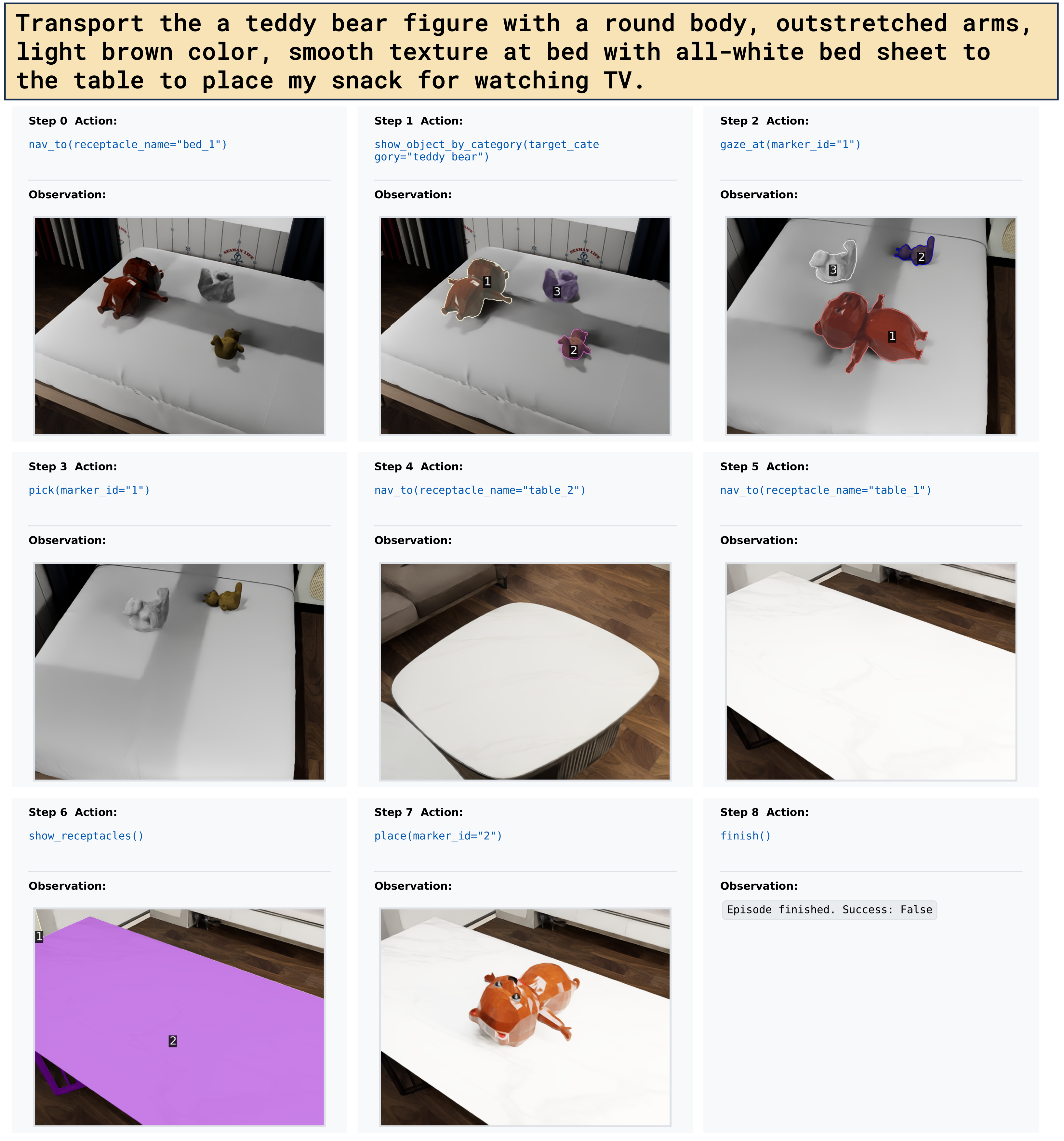}
  \caption{\textbf{Failure: Similar furniture confusion.} Task: move a teddy bear from a bed to the table for watching TV. The agent correctly picks the target object (Steps~0--3) but navigates between two candidate tables (Steps~4--5) and ultimately places it at the wrong destination (Step~7), yielding failure.}
  \label{fig:case_wrong_dest}
\end{figure}

\paragraph{Similar Furniture Confusion.}
\Cref{fig:case_wrong_dest} illustrates a task with the instruction: \textit{``Transport a teddy bear figure with a round body, outstretched arms, light brown color, and smooth texture from the bed with all-white bed sheets to the table for placing my snack while watching TV.''}
The agent navigates to \texttt{bed\_1} (Step~0), detects three teddy bears via \texttt{show\_object\_by\_category} (Step~1, markers~1--3), approaches marker~1 with \texttt{gaze\_at} (Step~2), and successfully picks it (Step~3).
During the placement phase, the agent navigates to \texttt{table\_2} (Step~4), which appears as an empty white surface. The intended destination is \texttt{table\_2} because it is the table in front of the TV and in the living room. However, the agent fails to semantically ground the destination description and then proceeds to \texttt{table\_1} (Step~5). It calls \texttt{show\_receptacles} to label nearby furniture (Step~6), revealing two receptacle markers. The agent places the teddy bear at marker~2 (Step~7), but this corresponds to the wrong table.
The destination description \textit{``the table to place my snack for watching TV''} requires commonsense inference that the intended target is the table near the TV area, a level of semantic grounding that the frozen vision encoder does not reliably capture.

\begin{figure}[!ht]
  \centering
  \includegraphics[width=\linewidth]{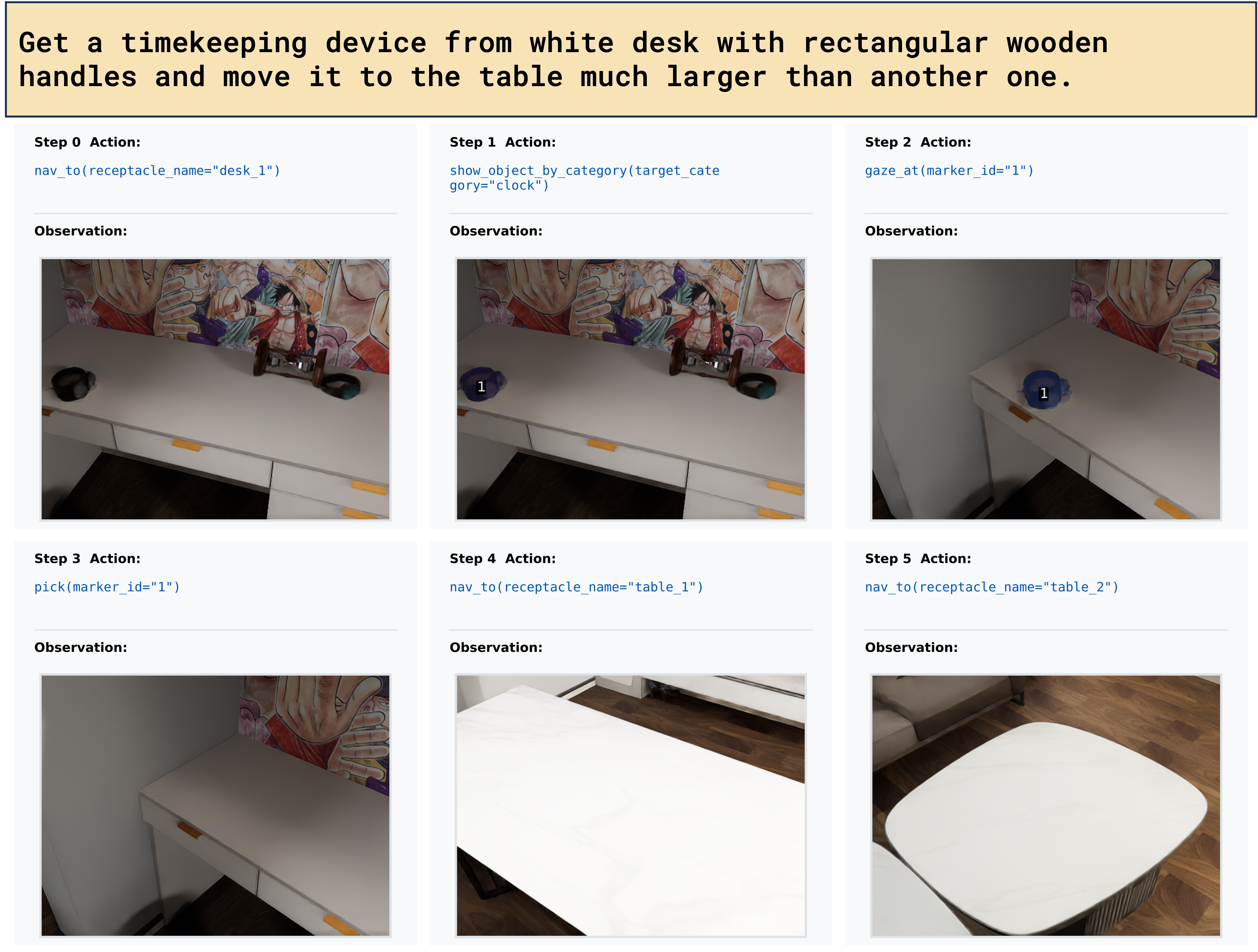}
  \caption{\textbf{Failure: Key action missing---no \texttt{ask}.} Task: retrieve a \textit{``timekeeping device''} from a desk with multiple candidate objects. The vague instruction creates ambiguity about which object to pick, but the agent selects one arbitrarily (Steps~1--3) without invoking \texttt{ask} to disambiguate, resulting in picking the wrong object.}
  \label{fig:case_no_ask}
\end{figure}

\paragraph{Key Action Missing.}
\Cref{fig:case_no_ask} shows a task with the instruction: \textit{``Get a timekeeping device from white desk with rectangular wooden handles and move it to the table much larger than another one.''}
The agent navigates to \texttt{desk\_1} (Step~0) and invokes \path{show_object_by_category(target_category="clock")} (Step~1), which reveals multiple candidate objects on the desk. The instruction uses the vague term \textit{``timekeeping device''} rather than specifying a particular object, creating genuine ambiguity about which candidate to retrieve. Instead of invoking the \texttt{ask} tool to query the simulated user for clarification, the agent arbitrarily gazes at one candidate (Step~2) and picks it (Step~3), selecting the wrong object and causing the task to fail.
This failure belongs to the \emph{key action missing} category (17\% of failures). The ambiguous instruction requires communicative disambiguation that the agent cannot resolve from visual observation alone. Despite RL training with social rewards to encourage proactive querying, the agent defaults to immediate action rather than seeking clarification when faced with underspecified instructions.

\begin{figure}[!ht]
  \centering
  \includegraphics[width=\linewidth]{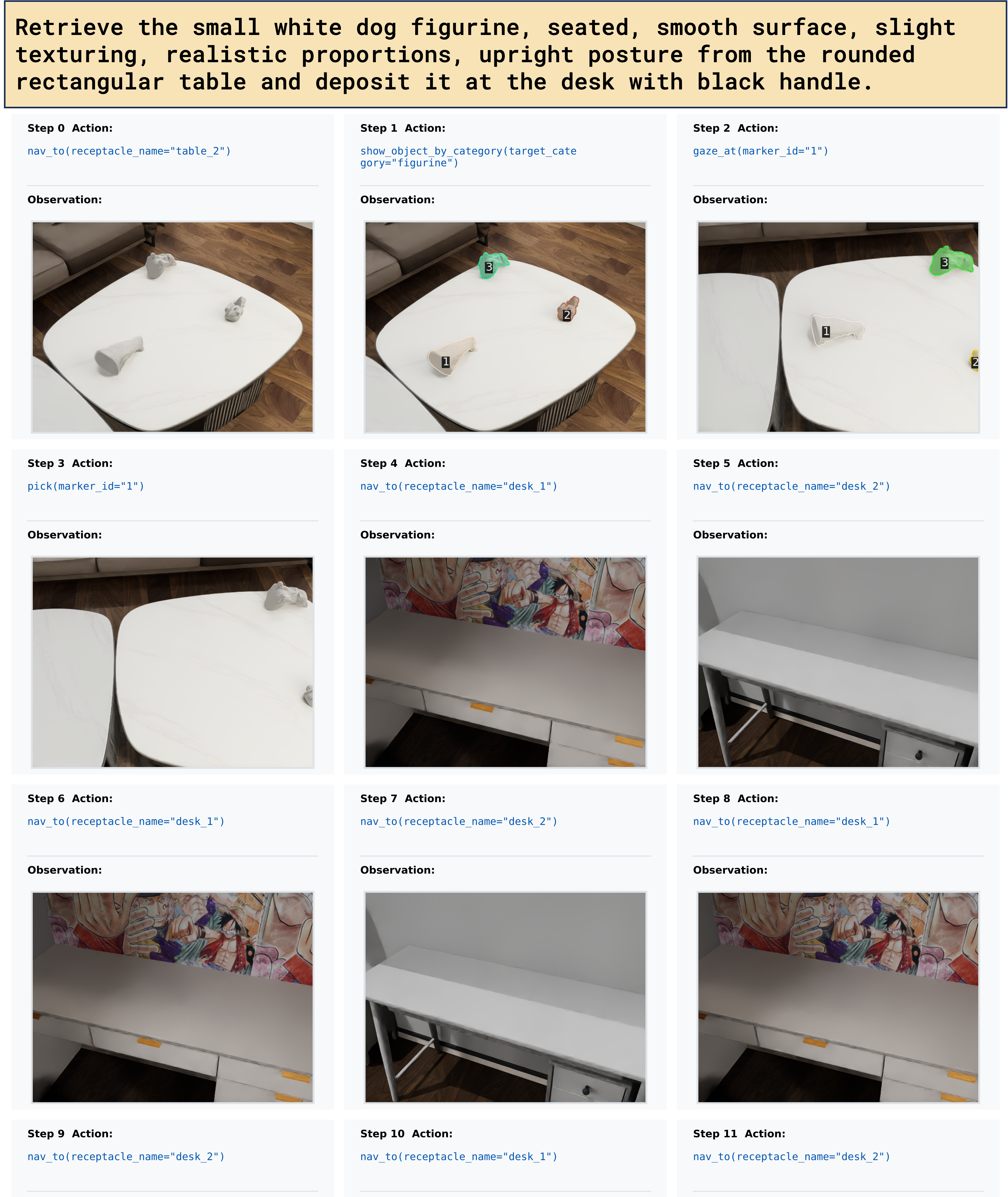}
  \caption{\textbf{Failure: Lost action memory---infinite navigation loop.} Task: deliver a dog figurine to \textit{``the desk with black handle.''} After picking the figurine (Steps~0--3), the agent alternates between \texttt{desk\_1} and \texttt{desk\_2} for eight steps (Steps~4--11) without placing the object or invoking any non-navigation tool.}
  \label{fig:case_infinite_loop}
\end{figure}

\paragraph{Lost Action Memory.}
\Cref{fig:case_infinite_loop} presents a long-horizon failure with the instruction: \textit{``Retrieve the small white dog figurine, seated, smooth surface, slight texturing, realistic proportions, upright posture from the rounded rectangular table and deposit it at the desk with black handle.''}
The pick phase completes without issue: the agent navigates to \texttt{table\_2} (Step~0), detects figurines via \path{show_object_by_category(target_category="figurine")} (Step~1), gazes at marker~1 (Step~2), and picks it (Step~3).
However, during the placement phase the agent enters a persistent oscillatory loop: it navigates to \texttt{desk\_1} (Step~4), then \texttt{desk\_2} (Step~5), then back to \texttt{desk\_1} (Step~6), and continues alternating between the two desks through Step~11---eight consecutive \texttt{nav\_to} calls without ever invoking \texttt{show\_receptacles}, \texttt{place}, or any other tool.
This behavior instantiates the \emph{lost action memory} failure mode: the self-summarization mechanism maintains a compact context by design, but the compressed history fails to encode that both desks have already been visited and visually inspected. Without this negative evidence, the agent cannot eliminate candidates or break the decision deadlock, producing an infinite loop that exhausts the step budget.

\subsection{Successful Exploration Cases}

We present success cases that highlight the emergent abilities of our agent. 

\begin{figure}[!ht]
  \centering
  \includegraphics[width=\linewidth]{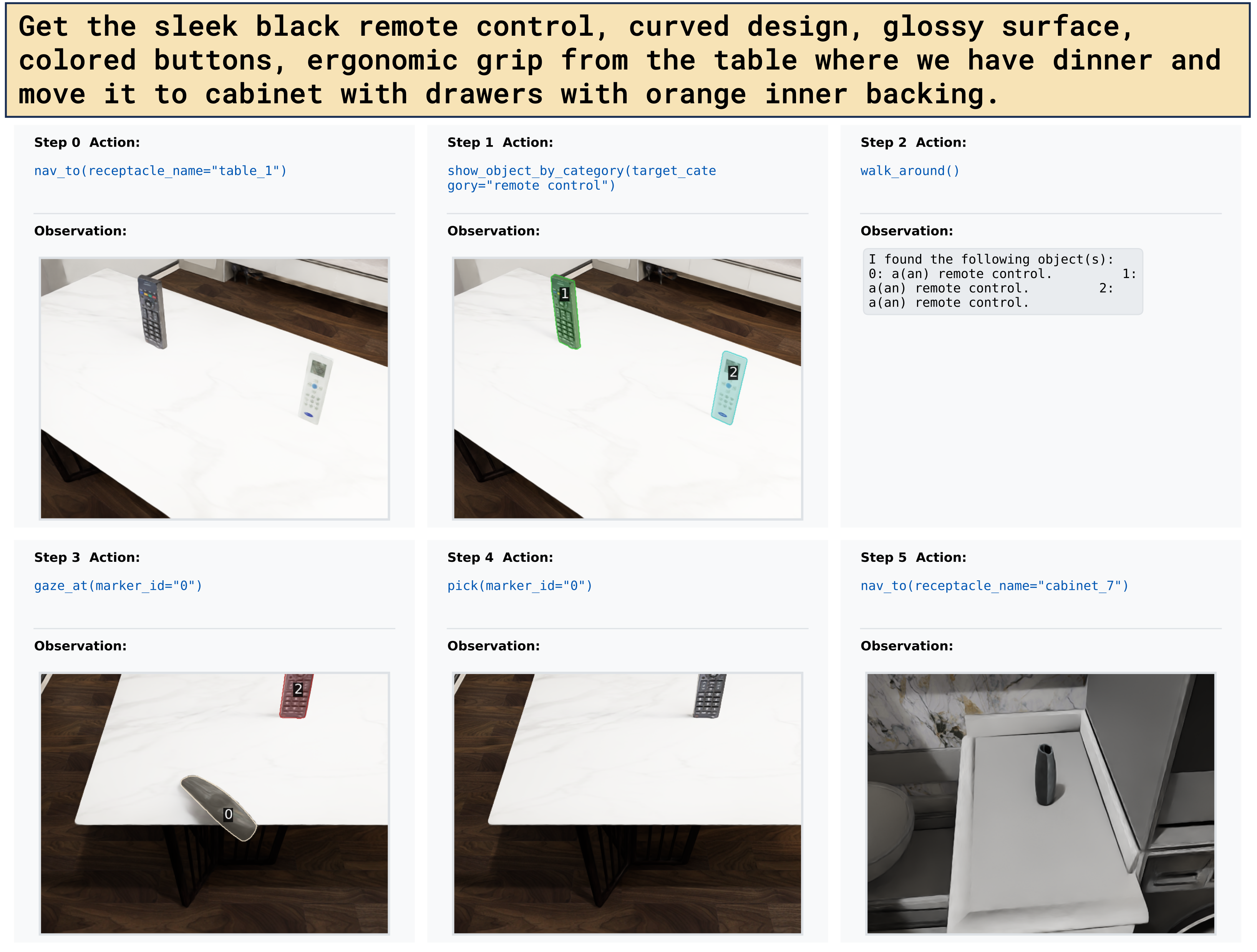}
  \caption{\textbf{Success: Active object discovery via \texttt{walk\_around}.} Task: retrieve a black remote control from a table with three differently-colored remotes. The agent detects all three candidates (Step~1), uses \texttt{walk\_around} to confirm the full inventory (Step~2), then correctly selects and picks the black remote (Steps~3--4) before delivering it to the target cabinet (Step~5).}
  \label{fig:case_walkaround}
\end{figure}

\paragraph{Active Object Discovery via \texttt{walk\_around}.}
\Cref{fig:case_walkaround} shows a task: \textit{``Get the sleek black remote control, curved design, glossy surface, colored buttons, ergonomic grip from the table where we have dinner and move it to cabinet with drawers with orange inner backing.''}
The agent navigates to \texttt{table\_1} (Step~0) and calls \texttt{show\_object\_\allowbreak by\_category} to search for remote controls (Step~1), revealing two visually distinct candidates annotated with markers~1 and~2, respectively.
Rather than immediately selecting one, the agent invokes \texttt{walk\_around()} (Step~2). The scan confirms the presence of all three candidates: \textit{``I found the following object(s): 0: a(an) remote control. 1: a(an) remote control. 2: a(an) remote control.''}
Having verified the inventory, the agent gazes at marker~0 (Step~3)---the sleek black remote matching the instruction---picks it (Step~4), and navigates to \texttt{cabinet\_7} (Step~5) to complete the task.
This episode illustrates the open-vocabulary exploration capability of our agent and the necessity of the \texttt{walk\_around} tool for comprehensive scene exploration.
The agent can use the multi-viewpoint \texttt{walk\_around} tool to discover objects that are invisible from the initial viewpoint.

\begin{figure}[!ht]
  \centering
  \includegraphics[width=\linewidth]{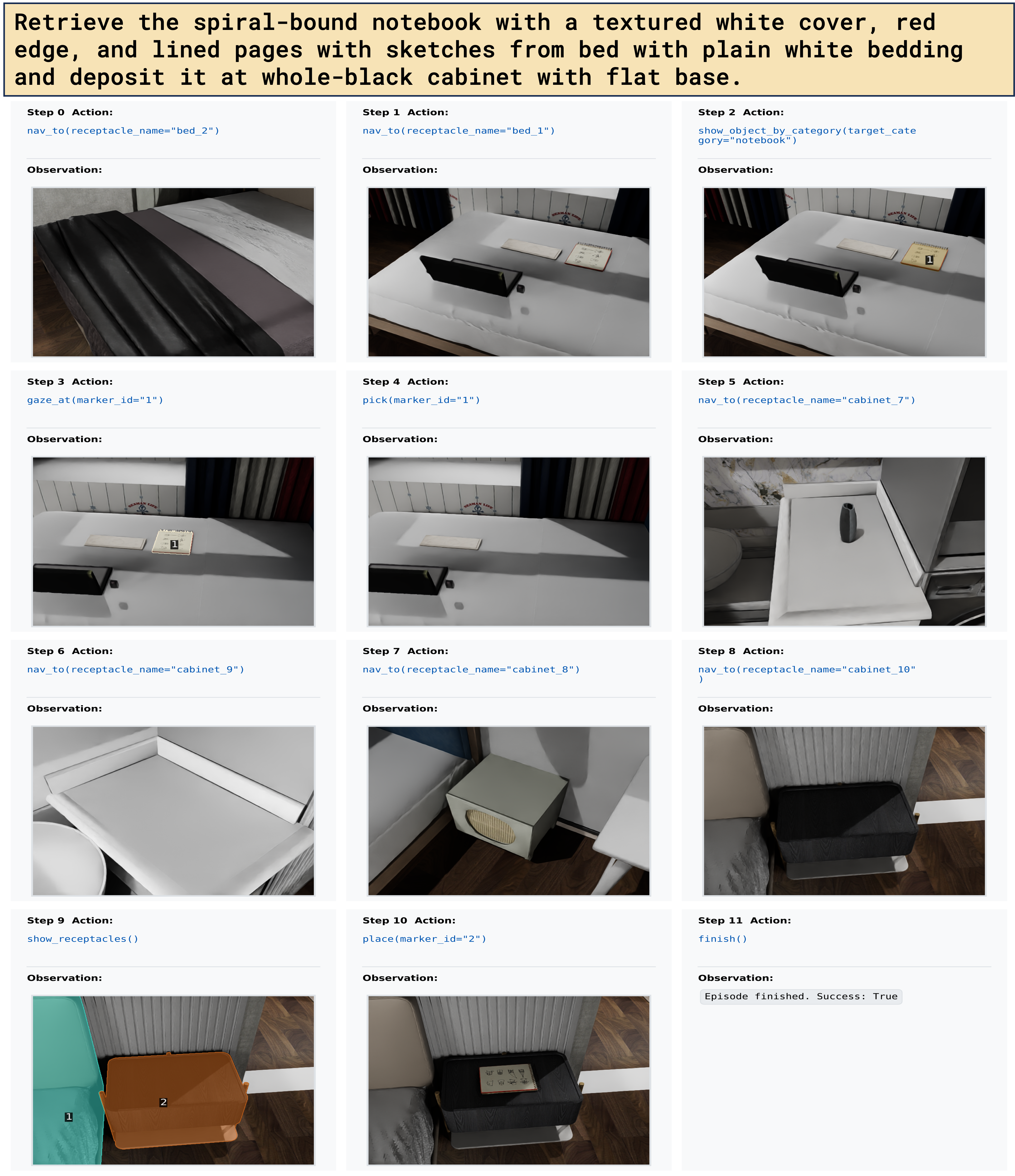}
  \caption{\textbf{Success: Multi-receptacle exploration task.} Task: move a notebook from \textit{``bed with plain white bedding''} to a \textit{``whole-black cabinet.''} The agent explores two beds to locate the correct source (Steps~0--1) and four cabinets to identify the correct destination (Steps~5--8), completing the 12-step task without revisiting any receptacle.}
  \label{fig:case_explore}
\end{figure}

\paragraph{Multi-Receptacle Exploration.}
\Cref{fig:case_explore} presents a task with the instruction: \textit{``Retrieve the spiral-bound notebook with a textured white cover, red edge, and lined pages with sketches from bed with plain white bedding and deposit it at whole-black cabinet with flat base.''}
Both the source and destination require visual grounding, as the scene contains multiple beds and multiple cabinets.
The agent first navigates to \texttt{bed\_2} (Step~0) but observes that the bedding does not match the description. It redirects to \texttt{bed\_1} (Step~1), where it detects the notebook via \texttt{show\_object\_by\_category (target\_category="notebook")} (Step~2), gazes at marker~1 (Step~3), and picks it (Step~4).
During the placement phase, the agent systematically explores four cabinets: \texttt{cabinet\_7} (Step~5), \texttt{cabinet\_9} (Step~6), \texttt{cabinet\_8} (Step~7), and \texttt{cabinet\_10} (Step~8). Upon reaching the last candidate, it invokes \texttt{show\_receptacles} (Step~9) to label visible furniture and confirm the correct destination, places the notebook at marker~2 (Step~10), and signals task completion with \texttt{finish()} (Step~11, \textit{``Success: True''}).
This trajectory demonstrates the agent's ability to execute systematic search across multiple candidate receptacles while maintaining goal persistence.
Unlike the infinite-loop failure mode, the agent here explores progressively, never revisiting a previously rejected candidate, and commits to a placement once it identifies the matching furniture.

  \FloatBarrier
  \setcounter{figure}{0}%
  \setcounter{table}{0}%
  \setcounter{algorithm}{0}%
  \section{Real-World Deployment Details}
\label{sec:supp_real_world}

To validate the effectiveness of our training pipeline in physical environments, we deploy our \realagent on a real-world mobile manipulator. This section details the hardware setup, the software architecture bridging the VLM to physical execution, and a qualitative walkthrough of a complex long-horizon task.

\subsection{Hardware and System Architecture}
\label{subsec:supp_real_world_system}

\paragraph{Hardware Platform.}
Real-world experiments are conducted on the \textbf{ARX LIFT2}, a highly capable dual-arm mobile manipulator. The robot is built upon a three-wheel omnidirectional chassis for agile indoor mobility and features a vertically adjustable torso that freely elevates and lowers the dual-arm platform to expand the operational workspace. For visual perception, it is equipped with three Intel RealSense cameras: one D455 mounted on the head for ego-centric global observations, and two D405 cameras on the wrists for high-resolution, close-up manipulation views. Autonomous navigation is powered by an Odin module on the chassis, which fuses LiDAR and RGB camera data. The robot features an onboard NVIDIA RTX 4070 Ti GPU that runs the low-level controllers and the MCP server in real time.

\paragraph{System Architecture and MCP Server.}
To seamlessly connect the high-level cognitive reasoning of the VLM with the low-level hardware controllers, we use the MCP server architecture (as introduced in \cref{sec:supp_tool_impl_per}). The VLM policy is deployed on a cloud server for high-performance inference. It takes the head camera's RGB image and the accumulated history as input and outputs discrete, JSON-formatted tool calls over the network to the robot. 

The onboard MCP server acts as the central dispatcher. When the cloud-based VLM issues a navigation command (\eg \texttt{nav\_to}), the MCP server routes the target coordinates to the Odin navigation module. When a manipulation command is issued (\eg \texttt{pick}, \texttt{place}, \texttt{open}, \texttt{close}), the MCP server passes the visual marker ID and the command to a cloud-deployed $\pi_{0.5}$ model running on an RTX 4090. To enable robust physical execution, we collected 980 real-world teleoperation trajectories to fine-tune this Pi-0.5 model, allowing it to reliably translate the wrist cameras' RGB-D inputs into continuous motor trajectories for the dual arms. If the VLM issues a communicative command (\eg \texttt{ask}), the MCP server triggers a text-to-speech/speech-to-text interface to interact with the human user. This modular design ensures that the tool paradigm, which is free of oracle perceptual APIs in simulation, transfers directly to the real world without modification.

\subsection{Quantitative Evaluation and Failure Analysis}
\label{subsec:supp_real_world_quant}

\paragraph{Quantitative Real-World Evaluation.}
We evaluate the physical deployment at three levels: repeatability of navigation and gaze alignment, executability of low-level VLA primitives, and end-to-end task success. As shown in \cref{tab:supp_real_world_quant}, the robot achieves centimeter-level navigation/gaze repeatability over 30 trials, 85.3\% primitive success over 600 executions, and 78.3\% full-system success over 60 episodes, with no unrecoverable system crashes.
The \emph{VLM Lat.} column reports only the cumulative high-level VLM inference time during an episode. Low-level physical execution has a separate fixed horizon: each Pi-0.5 manipulation primitive runs for 1000 low-level control steps, grouped into 50-step action chunks, and takes approximately 60\,s on the physical platform. Thus, the dominant wall-clock deployment cost comes from physical robot execution rather than VLM inference.

\begin{table}[t]
\centering
% \scriptsize
% \setlength{\tabcolsep}{3pt}
% \renewcommand{\arraystretch}{0.92}
\caption{\textbf{Quantitative real-world evaluation.} Odom Err., Yaw Err., and Img. Shift measure repeatability after repeated navigation/gaze commands. Primitive VLA executability is measured over 600 low-level executions. End-to-end episodes use the same high-level policy and MCP tool interface as simulation.}
\label{tab:supp_real_world_quant}
\begin{tabular}{@{}lccccc@{}}
\toprule
\multicolumn{6}{c}{\textit{Navigation/gaze repeatability}} \\
\midrule
Target & N trials & Odom Err. & Yaw Err. & Img. Shift & RMSE \\
\midrule
All & 30 & $1.78{\pm}0.40$\,cm & $0.74{\pm}0.57^\circ$ & $21.5{\pm}10.4$\,px & 1.96\,px \\
\midrule
\multicolumn{6}{c}{\textit{Primitive VLA executability}} \\
\midrule
\textbf{Skill} & N & SR & \textbf{Skill} & N & SR \\
\midrule
\textbf{Open} & 120 & 63.3\% & \textbf{Close} & 120 & 100.0\% \\
\textbf{Pick} & 120 & 89.2\% & \textbf{Place-desk} & 120 & 100.0\% \\
\textbf{Place-micro.} & 120 & 74.2\% & \textbf{Overall} & 600 & 85.3\% \\
\midrule
\multicolumn{6}{c}{\textit{End-to-end real-world episodes}} \\
\midrule
Task & SR & Steps & SPL & Ask & VLM Lat. \\
\midrule
FDO & 80.0\% & 12.2 & 67.0\% & 0.90 & 71.1s \\
FODP & 100.0\% & 9.6 & 86.9\% & 0.70 & 53.5s \\
SUL & 55.0\% & 13.0 & 35.4\% & 0.55 & 80.5s \\
\textbf{Overall} & \textbf{78.3\%} & \textbf{11.6} & \textbf{63.1\%} & \textbf{0.72} & \textbf{68.4s} \\
\bottomrule
\end{tabular}
\end{table}

\paragraph{Real-World Failure Breakdown.}
Among the 60 end-to-end physical episodes, only 2 episodes failed because of high-level visual misunderstanding by the frozen vision encoder. The remaining failures are caused by low-level VLA execution errors, physical timeouts, or missing \texttt{ask} calls under ambiguity. This breakdown indicates that the deployed high-level policy transfers its visual-interactive reasoning to the physical setting, while overall robustness still depends on improving primitive reliability and timely user-query behavior.

\subsection{Qualitative Execution Walkthrough}
\label{subsec:supp_real_world_walkthrough}

\paragraph{Qualitative Execution Walkthrough (12-Step Task).}
As highlighted in the main text (\cref{sec:experiments}), we evaluate the agent on a complex, semantically ambiguous task: fetching a specific food item from a bread stand and placing it into a microwave. The execution spans 12 distinct steps, demonstrating the agent's capabilities:

\begin{enumerate}
    \item \textbf{Instruction Reception \& Predictive Planning:} The agent receives the user's high-level instruction. Instead of immediately searching for the food, the agent exhibits causal reasoning by first navigating to the microwave.
    \item \textbf{Preparation (Open):} The agent executes the \texttt{open} tool to open the microwave door, preparing the destination receptacle in advance.
    \item \textbf{Navigation to Source:} The agent navigates across the room to the bread stand.
    \item \textbf{Active Exploration:} The agent invokes the visual perception tool to scan the bread stand, generating SoM overlays on the observed items.
    \item \textbf{Ambiguity Detection:} The agent processes the visual observation and realizes there are multiple visually similar food items (distractors) that match the vague instruction.
    \item \textbf{Human-Robot Disambiguation (Ask):} Rather than guessing, the agent proactively halts physical execution and uses the \texttt{ask} tool to query the user for clarification regarding the specific target.
    \item \textbf{User Feedback:} The user provides a clarifying response (\eg specifying the color or exact location of the target item).
    \item \textbf{Target Grounding:} The agent integrates the user's feedback with the current visual observation to uniquely identify the correct target ID.
    \item \textbf{Precise Manipulation (Pick):} The agent executes the \texttt{pick} tool, successfully grasping the disambiguated food item.
    \item \textbf{Return Navigation:} The agent navigates back to the microwave with the item in hand.
    \item \textbf{Task Execution (Place):} The agent executes the \texttt{place} tool, carefully depositing the food item inside the open microwave.
    \item \textbf{Task Completion:} The agent concludes the episode, successfully completing the long-horizon, closed-loop task.
\end{enumerate}

This 12-step trajectory validates that the RL-optimized policy (GSPO) effectively overcomes distribution shifts, enabling adaptive planning, active visual grounding, and proactive social interaction in unstructured physical environments.

\begin{figure}[htp!]
    \centering
    \includegraphics[width=0.8\linewidth]{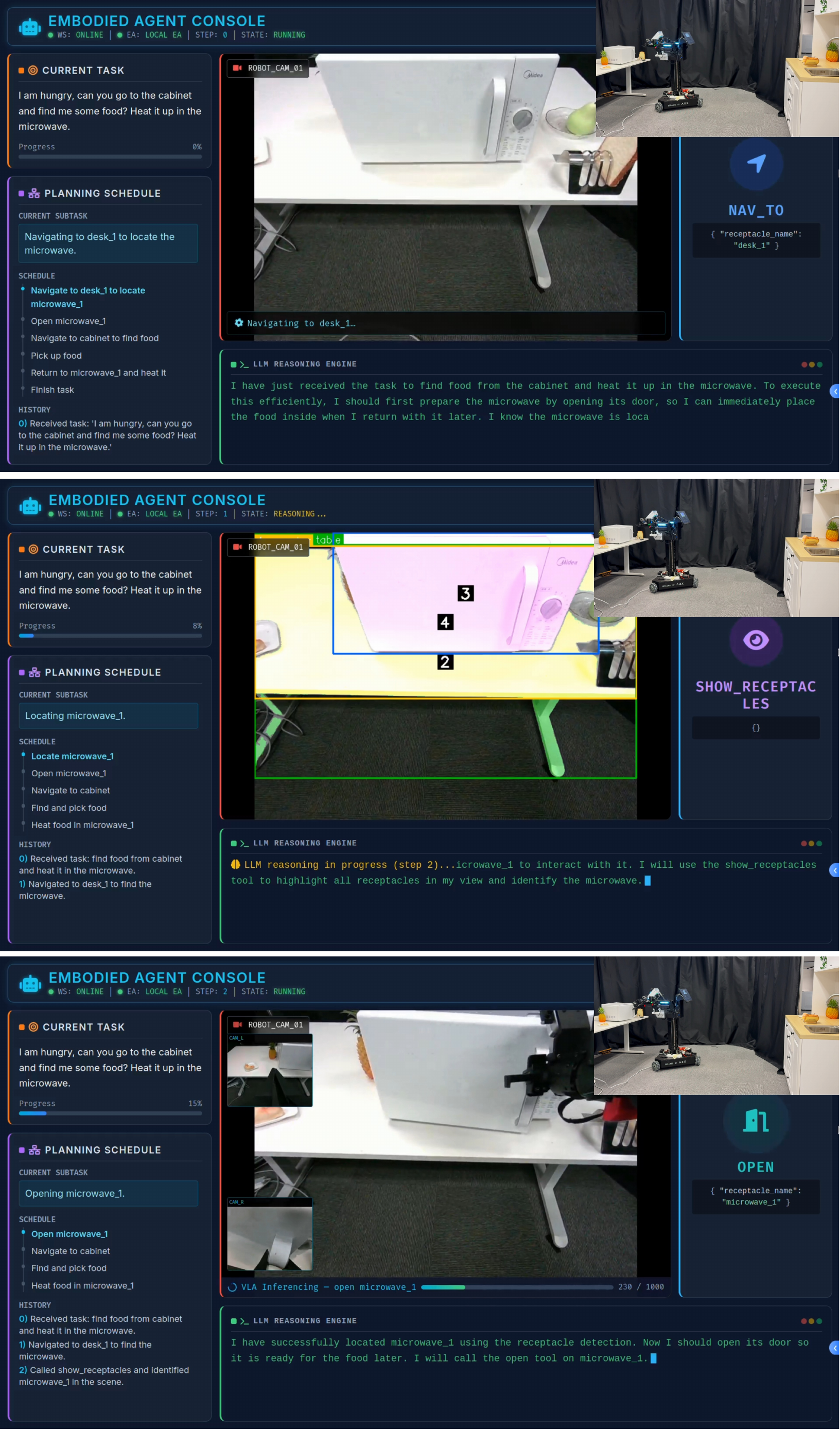}
    \caption{Execution sequence of the real-world task (Steps~1--3).}
    \label{fig:real_1_3}
\end{figure}

\begin{figure}[htp!]
    \centering
    \includegraphics[width=0.8\linewidth]{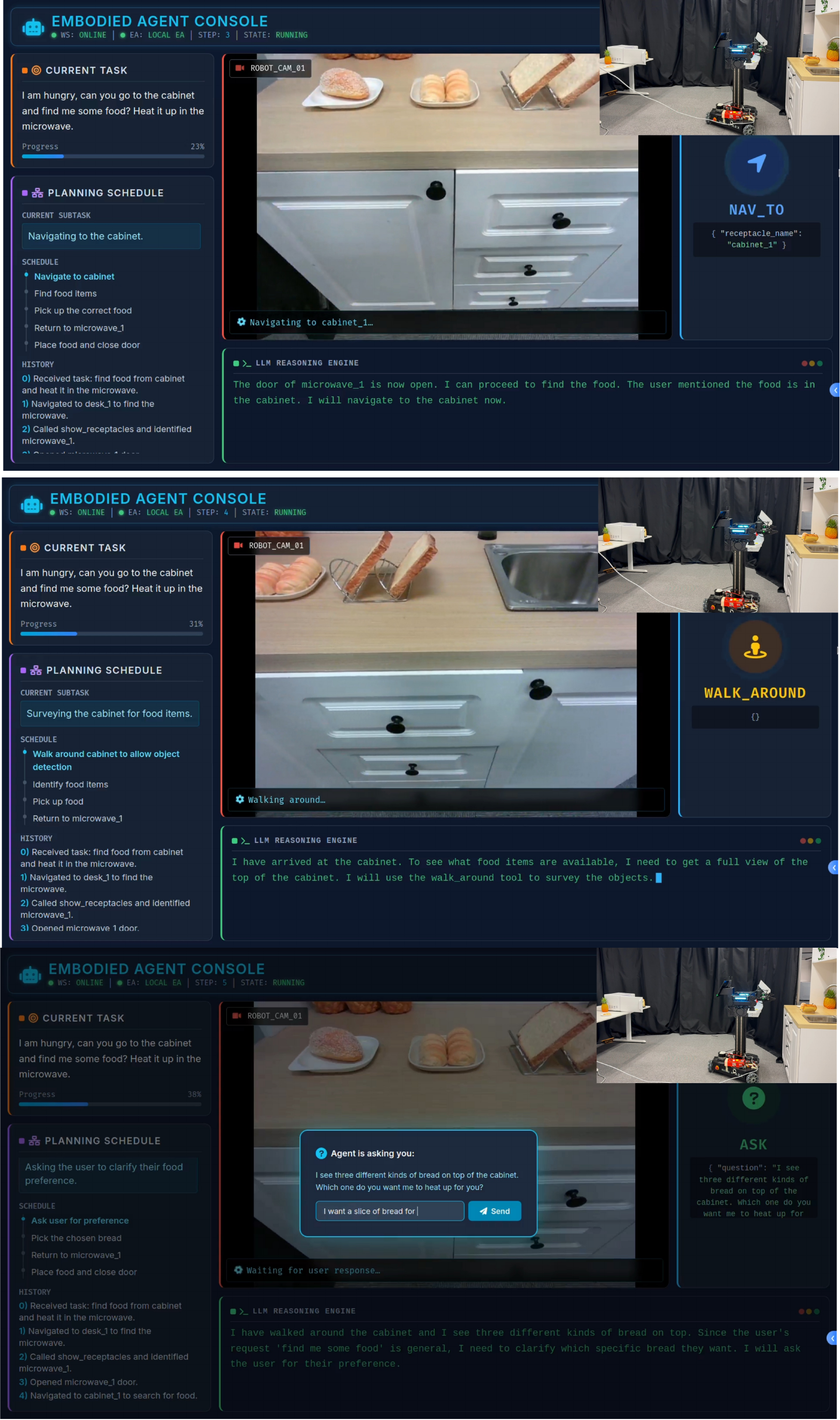}
    \caption{Execution sequence of the real-world task (Steps~4--6).}
    \label{fig:real_4_6}
\end{figure}

\begin{figure}[htp!]
    \centering
    \includegraphics[width=0.8\linewidth]{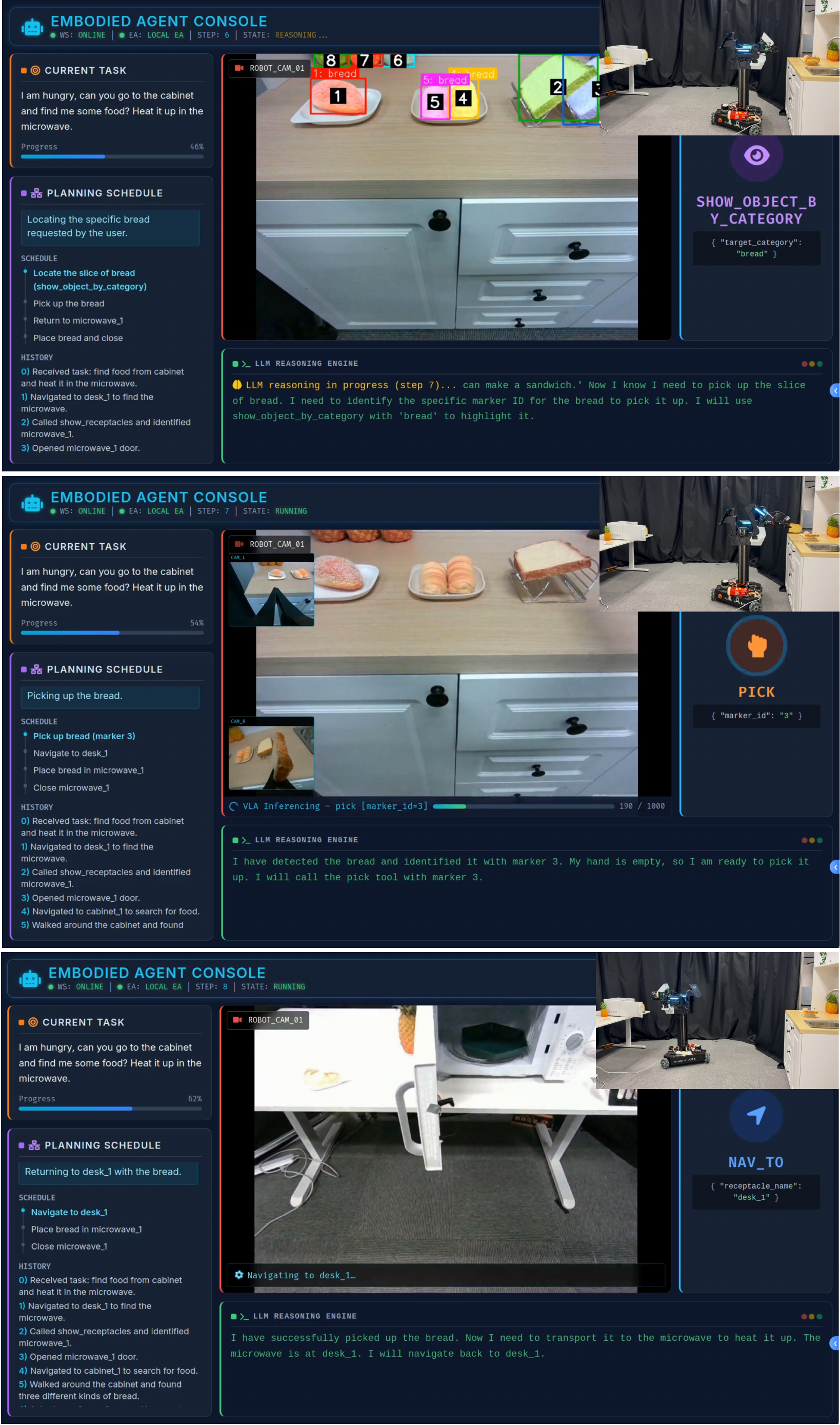}
    \caption{Execution sequence of the real-world task (Steps~7--9).}
    \label{fig:real_7_9}
\end{figure}

\begin{figure}[htp!]
    \centering
    \includegraphics[width=0.8\linewidth]{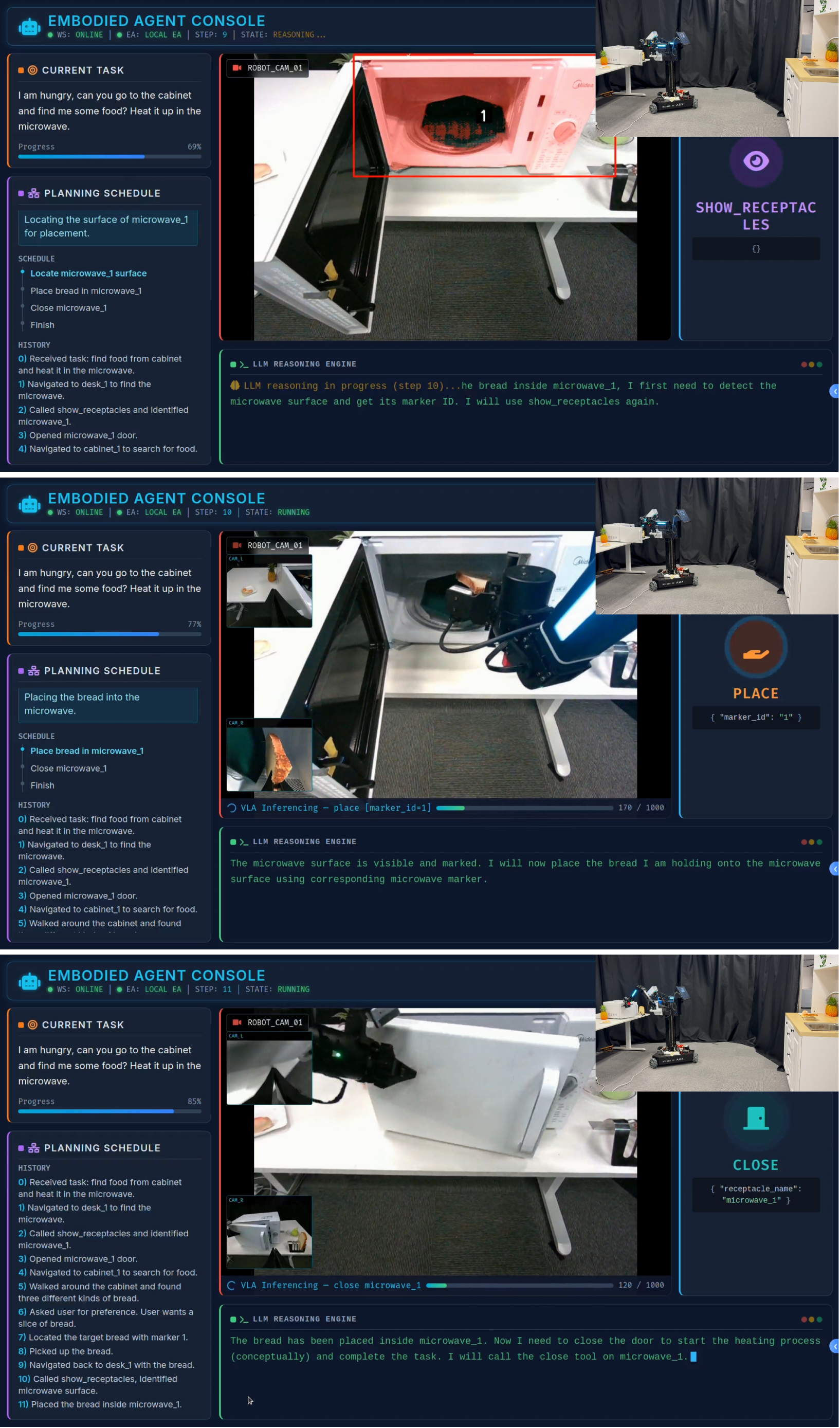}
    \caption{Execution sequence of the real-world task (Steps~10--12).}
    \label{fig:real_10_12}
\end{figure}

The progressive execution of the real-world task, from Step~1 to Step~12, is illustrated in sequence in Figures~\ref{fig:real_1_3}--\ref{fig:real_10_12}.

% --- 第 1 张图 (Steps 1-3) ---

\clearpage % 强制当前页的内容排版完毕，确保下一张图从新的一页开始

% --- 第 2 张图 (Steps 4-6) ---

\clearpage

% --- 第 3 张图 (Steps 7-9) ---

\clearpage

% --- 第 4 张图 (Steps 10-12) ---

\FloatBarrier

\bibliography{main}

\end{document}